\documentclass{article}
\usepackage[preprint]{colm2026_conference}
\usepackage{enumitem}
\usepackage{microtype}
\usepackage{hyperref}
\usepackage{url}
\usepackage{booktabs}
\usepackage{multirow}
\usepackage{colortbl}
\usepackage{makecell}
\usepackage{graphicx}
\usepackage{amsmath}
\usepackage{amssymb}
\usepackage{lineno}
\usepackage{xcolor}
\usepackage{float}
\usepackage{placeins}
\usepackage{wrapfig}
\usepackage{subcaption}
\usepackage{graphicx}

\definecolor{SoftGreen}{RGB}{60,122,78}
\definecolor{SoftRed}{RGB}{196,96,96}

\usepackage[most]{tcolorbox}
\usepackage{xcolor}

\usepackage[most]{tcolorbox}
\usepackage{xcolor}

\definecolor{boxgreen}{HTML}{4A7C59}
\definecolor{bggreen}{HTML}{E8F5E9}
\definecolor{boxblue}{HTML}{3B5998}
\definecolor{bgblue}{HTML}{E3EDF9}
\definecolor{boxteal}{HTML}{2E7D7B}
\definecolor{bgteal}{HTML}{E0F2F1}
\definecolor{boxamber}{HTML}{8D6E2C}
\definecolor{bgamber}{HTML}{FFF8E1}

\definecolor{bghold}{HTML}{DDE4EC}
\definecolor{bgcorrect}{HTML}{C6E7D0}
\definecolor{bgincorrect}{HTML}{F0C8C8}

\newtcolorbox{promptbox}[1][]{
  enhanced, breakable,
  colback=bggreen, colframe=boxgreen, boxrule=1.0pt, arc=3pt,
  left=8pt, right=8pt, top=8pt, bottom=8pt,
  fonttitle=\bfseries, coltitle=white, colbacktitle=boxgreen,
  attach boxed title to top center={yshift=-2mm},
  title=#1
}

\newtcolorbox{promptboxblue}[1][]{
  enhanced, breakable,
  colback=bgblue, colframe=boxblue, boxrule=1.0pt, arc=3pt,
  left=8pt, right=8pt, top=8pt, bottom=8pt,
  fonttitle=\bfseries, coltitle=white, colbacktitle=boxblue,
  attach boxed title to top center={yshift=-2mm},
  title=#1
}

\newtcolorbox{promptboxteal}[1][]{
  enhanced, breakable,
  colback=bgteal, colframe=boxteal, boxrule=1.0pt, arc=3pt,
  left=8pt, right=8pt, top=8pt, bottom=8pt,
  fonttitle=\bfseries, coltitle=white, colbacktitle=boxteal,
  attach boxed title to top center={yshift=-2mm},
  title=#1
}

\newtcolorbox{promptboxamber}[1][]{
  enhanced, breakable,
  colback=bgamber, colframe=boxamber, boxrule=1.0pt, arc=3pt,
  left=8pt, right=8pt, top=8pt, bottom=8pt,
  fonttitle=\bfseries, coltitle=white, colbacktitle=boxamber,
  attach boxed title to top center={yshift=-2mm},
  title=#1
}

\definecolor{darkblue}{rgb}{0,0,0.5}
\hypersetup{colorlinks=true,citecolor=darkblue,linkcolor=darkblue,urlcolor=darkblue}

\title{Benchmarking Multi-turn Medical Diagnosis: \\Hold, Lure, and Self-Correction}

\author{
\\
\begin{tabular}{c}
\normalfont
\textbf{Jinrui Fang}$^{1}$, \textbf{Runhan Chen}$^{2}$, \textbf{Xu Yang}$^{1}$, \textbf{Jian Yu}$^{1}$, \textbf{Jiawei Xu}$^{1}$, \\
\textbf{Ashwin Vinod}$^{1}$, \textbf{Wenqi Shi}$^{3}$, \textbf{Tianlong Chen}$^{4}$, \textbf{Heng Ji}$^{5}$, \textbf{ChengXiang Zhai}$^{5}$, \\
\textbf{Ying Ding}$^{1}$\thanks{Corresponding authors.}, \textbf{Yuji Zhang}$^{5}$\footnotemark[1] \\
[0.8em]
{\normalfont $^{1}$The University of Texas at Austin} \\
{\normalfont $^{2}$New York University} \\
{\normalfont $^{3}$The University of Texas Southwestern Medical Center} \\
{\normalfont $^{4}$University of North Carolina at Chapel Hill} \\
{\normalfont $^{5}$University of Illinois Urbana-Champaign} \\
[0.4em]
{\normalfont \texttt{jinrui@utexas.edu, ying.ding@ischool.utexas.edu, yujiz@illinois.edu}}
\end{tabular}
}

\begin{document}

\ifcolmsubmission
\linenumbers
\fi

\maketitle

\NewDocumentCommand{\yuji}
{ mO{} }{\textcolor{blue}{\textsuperscript{{Yuji}}{{\small[#1]}}}}

\NewDocumentCommand{\jinrui}
{ mO{} }{\textcolor{orange}{\textsuperscript{{Jinrui}}{{\small[#1]}}}}

\NewDocumentCommand{\runhan}
{ mO{} }{\textcolor{purple}{\textsuperscript{{Runhan}}{{\small[#1]}}}}

\newcommand{\ws}[1]{\textcolor{blue}{WenqiS: #1}}

\begin{abstract}

Large language models (LLMs) achieve high accuracy in medical diagnosis when all clinical information is provided in a single turn, yet how they behave under multi-turn evidence accumulation closer to real clinical reasoning remains unexplored. We introduce \textbf{MINT}\footnotemark (\textbf{M}edical \textbf{I}ncremental \textbf{N}-\textbf{T}urn Benchmark), a high-fidelity, multi-turn medical diagnosis benchmark comprising 1,035 cases with clinically labeled evidence shards, controlled turn granularity, and information-preserving decomposition. Through systematic evaluation of 11 LLMs on MINT, we uncover three persistent behavioral patterns that significantly impact diagnostic decisions: (1) \emph{intent to answer}, models rush to answer before sufficient evidence has been observed, with over 55\% of answers committed within the first two turns; (2) \emph{self-correction}, incorrect-to-correct answer revisions occur at up to 10.6$\times$ the rate of correct-to-incorrect flips, revealing a latent capacity for self-correction that premature commitment forecloses; and (3) \emph{strong lures}, clinically salient information such as laboratory results trigger premature answering even when models are explicitly instructed to wait. We translate these findings into clinically actionable guidance: deferring the diagnostic question to later turns reduces premature answering and improves accuracy at the first point of commitment by up to 62.6\%, while reserving salient clinical evidence for later turns prevents a catastrophic accuracy drop of up to 23.3\% caused by premature commitment. Our work provides both a controlled evaluation framework and concrete recommendations for improving the reliability of LLMs in multi-turn medical diagnosis.

\footnotetext{The data and code will be released upon publication.}
\end{abstract}

\section{Introduction}
Large language models (LLMs) have demonstrated strong performance in medical diagnosis under standard single-turn evaluation, where all clinical information is presented at once~\citep{chen2025medbullets,wang2024direct}. However, real-world clinical reasoning rarely follows this static paradigm~\citep{liu2025interactive,fan2025aihospital,li2024mediqquestionaskingllmsbenchmark}. In practice, clinicians gather evidence sequentially through multi-turn interactions (e.g., history taking, iterative diagnostic workup), gradually refining their judgment as new information emerges~\citep{liu2025interactive,fan2025aihospital,li2024mediqquestionaskingllmsbenchmark}. This discrepancy raises a critical question: can LLMs maintain reliable diagnostic reasoning in sequential, multi-turn settings?

Recent observations suggest that LLM performance degrades substantially when transitioning from single-turn to multi-turn settings \citep{laban2026lost,deshpande2025multichallenge}, particularly in high-stakes domains such as medicine \citep{liu2025interactive,fan2025aihospital,li2024mediqquestionaskingllmsbenchmark}. Several hypotheses have been proposed to explain this degradation. One line of reasoning attributes it to the model's tendency to commit prematurely to an answer upon receiving partial information, without sufficient reasoning \citep{laban2026lost,li2024mediqquestionaskingllmsbenchmark}. Another suggests that models lose track of relevant context over extended interactions \citep{laban2026lost,deshpande2025multichallenge,li2025structflowbench}. Despite these accounts, a fundamental research gap remains: \textbf{existing evaluations lack a controlled benchmark that can disentangle the effects of decision timing (\emph{hold}), restoration (\emph{self-correction}), and information exposure (\emph{lure}) on model performance.} Without such a benchmark, it remains unclear whether observed degradation stems from genuine model limitations or from artifacts introduced by uncontrolled data construction and evaluation protocols.

To address this gap, we construct MINT, a high-fidelity, fine-grained, multi-turn medical diagnosis benchmark that enables systematic investigation of model behavior under sequential evidence accumulation. MINT is built around three design principles: information preservation, controlled turn granularity, and clinically labeled evidence shards. Building on MINT, we conduct a comprehensive analysis of the mechanisms underlying multi-turn performance degradation and identify three persistent behavioral patterns: (1) \textbf{Intent to answer.} By controlling when diagnostic questions are presented, we observe a large gap between initial and final answer accuracy when questions are presented early, indicating that premature answering significantly harms performance.
Withholding the diagnostic question until all clinical information has been revealed largely restores accuracy to single-turn levels, demonstrating that models retain strong diagnostic potential when early answering is prevented. 
This result offers a key distinction from prior work, which has documented premature answering as a failure mode~\citep{li2024mediqquestionaskingllmsbenchmark, laban2026lost, luo2025clarifymt} but has not systematically isolated answer timing as a controlled variable: models retain strong diagnostic potential when commitment is deferred to later turns.
(2) \textbf{Self-correction.} We further analyze how holding intent shapes model reasoning dynamics across turns. Notably, transitions from incorrect to correct answers occur at a significantly higher rate (10.6$\times$) than transitions from correct to incorrect. This finding suggests that holding intent does not merely delay answering, but actively unlocks the model's capacity for self-correction during multi-turn reasoning. (3) \textbf{Strong lures.} Certain types of clinical information (e.g., lab results) act as strong lures that trigger immediate answering behavior. Models are significantly more likely to answer prematurely when laboratory results appear in early turns, an effect especially pronounced for lab-dependent diseases, leading to substantially higher error rates. These findings reveal that answering behavior is driven not only by internal intent, but also by the type and timing of incoming clinical evidence.

\begin{figure}[t]
    \centering
    \includegraphics[width=1\linewidth]{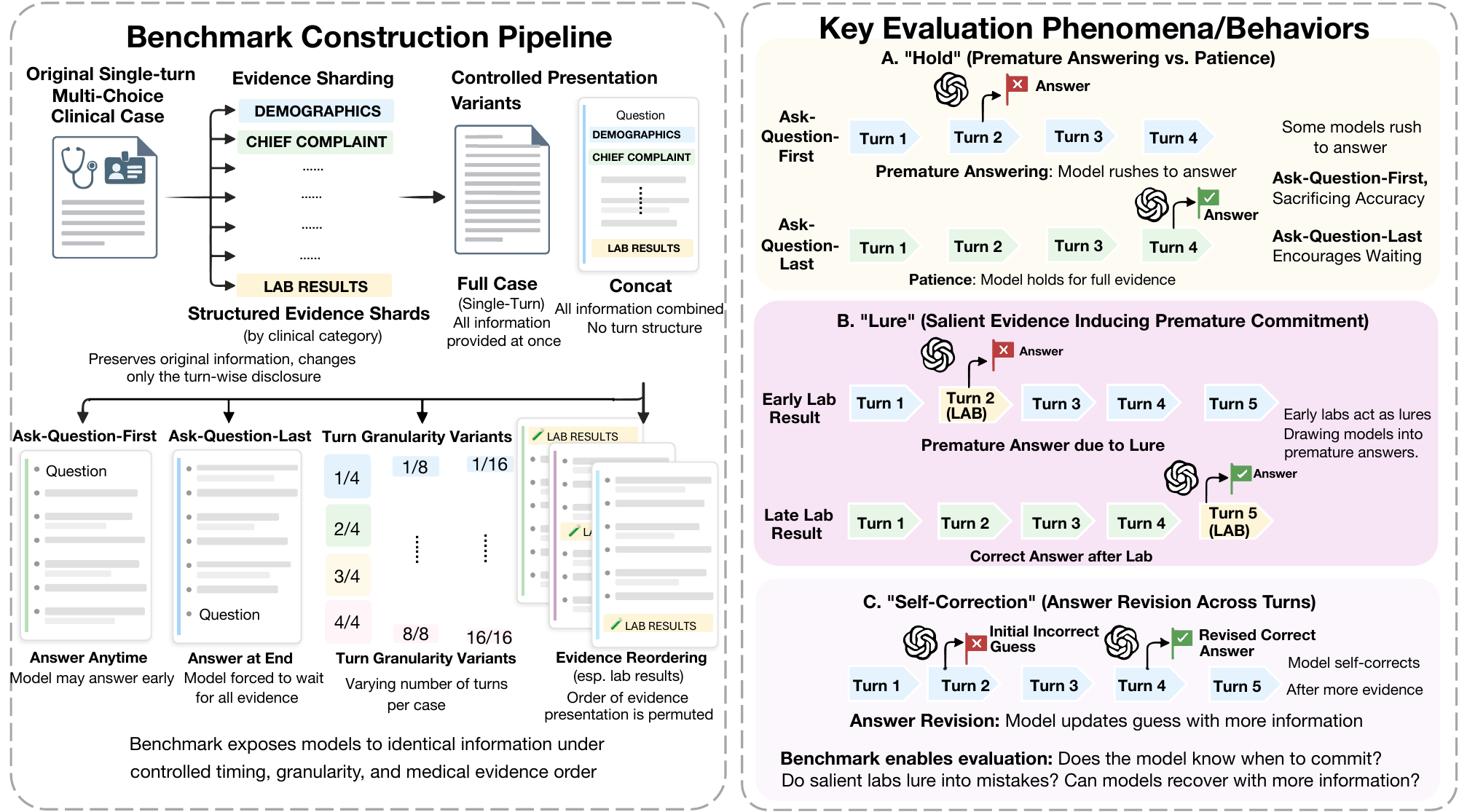}
    
    \caption{Overview of MINT benchmark construction and evaluation settings.}

    \label{fig:figure1}
\end{figure}

In summary, this work makes the following novel contributions (illustrated in Figure~\ref{fig:figure1}):
\begin{itemize}[leftmargin=*]
    \item \textbf{MINT Benchmark.} We construct a high-fidelity, fine-grained multi-turn medical diagnosis benchmark, MINT, with controlled decomposition, verification, and structured labeling, enabling rigorous study of LLM behavior in multi-turn medical diagnosis.
    \item \textbf{Intent to Answer.} By controlling when diagnostic questions are posed, we identify a universal pattern of premature commitment across all evaluated models and demonstrate that deferring the question to later turns substantially recovers diagnostic accuracy.
    \item \textbf{Self-Correction.} By tracking reasoning dynamics across turns, we demonstrate a persistent pattern of strong self-correction across all models, with commercial LLMs demonstrating up to 10.6$\times$ higher rates of turning incorrect answers into correct ones.
    \item \textbf{Strong Lures.} By presenting clinically salient information (e.g., lab results) in early turns, we uncover a consistent pattern in which strong lures reliably trigger premature diagnostic decisions, leading to substantially higher error rates.
\end{itemize}

\section{Related Work}
\textbf{Multi-turn Degradation in LLMs.} Prior work has consistently shown that LLMs are less reliable in multi-turn than in single-turn settings. Multi-turn competence varies widely across tasks \citep{deshpande2025multichallenge}, instruction-following degrades across turns~\citep{sun2024parrot}, and even frontier commercial models struggle with sustained contextual reasoning \citep{deshpande2025multichallenge,laban2026lost}. These failures often stem not from longer context per se, but from structured cross-turn dependencies, where models fail to preserve earlier constraints or reconcile conflicting instructions introduced at different turns \citep{laban2026lost}. Recent work further characterizes this as models getting ''lost'' in multi-turn conversations \citep{laban2026lost,graf2026turnwise,hankache2025evaluating}, suggesting that prior context alone can distort downstream behavior. However, these studies do not examine how diagnostic commitment unfolds under controlled sequential evidence. MINT brings this question into medicine by analyzing when models answer and how their predictions evolve across turns.

\textbf{Hold, Abstention, and Premature Answering.} Prior work has identified premature answering under incomplete information as a recurring failure mode in multi-turn interactions \citep{luo2025clarifymt,zhao2026and}, where LLMs often form early assumptions and become less reliable as later context arrives \citep{laban2026lost,liu2026intent}. This concern is especially consequential in medicine, where diagnosis unfolds through physician--patient dialogue, history taking, and adaptive questioning \citep{tu2025towards,johri2025evaluation,Werthaim2026Q4Dx}, making premature or overconfident answers particularly prone to unsafe or unreliable outcomes \citep{draelos2026large,griot2025large}. One line encourages models to continue gathering information by asking clarifying or follow-up questions rather than committing when evidence is insufficient \citep{li2024mediqquestionaskingllmsbenchmark,gatto2025follow,zhao2026and}. Another studies uncertainty-aware abstention \citep{machcha2025large,machcha2026knowing,dang2026knowguard}, enabling models to defer or withhold a response when available evidence does not yet support a reliable decision \citep{dang2026knowguard,watanabe2026clindet}. Recent work further shows that in clinical multi-turn settings, models may abandon correct early assessments in favor of later suggestions \citep{guo2026stop}. However, these studies primarily treat delayed answering as a defensive mechanism. It remains unclear whether withholding answers improves performance, whether self-correction emerges as evidence accumulates, and whether certain clinical evidence disproportionately triggers premature decisions.

\textbf{Multi-Turn Evaluation in Medicine.} Recent work has emphasized evaluating LLMs in multi-turn medical settings, moving beyond static question answering toward doctor--patient dialogue that more closely reflects clinical practice~\citep{johri2025evaluation,liao2024automatic,liu2025interactive,tu2025towards}. Yet current medical benchmarks remain insufficient for assessing clinical readiness in real-world settings~\citep{hager2024evaluation,johri2025evaluation,wang2025novel}. Prior work has therefore introduced interactive evaluation frameworks for multi-turn consultations, including state-aware patient simulators and task-oriented dialogue settings that emphasize proactive communication and overall consultation ability~\citep{liao2024automatic,liu2025interactive}. However, existing studies mainly evaluate consultation quality, communication ability, or overall diagnostic performance in simulated interactions~\citep{johri2025evaluation,liao2024automatic,liu2025interactive,tu2025towards}, rather than isolating how model predictions evolve when the same clinical case is revealed progressively under controlled conditions. In particular, they do not systematically control how clinical information is distributed across turns, verify information preservation after turn conversion, or manipulate the timing, granularity, and clinically salient type of evidence to study when models answer, revise, or continue to wait. To address this gap, MINT enables systematic study of how model predictions evolve under sequential evidence accumulation when the same clinical case is presented in a controlled, information-preserving manner, and how the timing, granularity, and type of clinical evidence jointly shape diagnostic behavior across turns.
\vspace{-0.5em}

\raggedbottom
\vbadness=10000
\vfuzz=30pt
\section{MINT: Benchmark Construction and Evaluation Setup}

\textbf{Data Sources.} 
We construct our multi-turn medical diagnosis benchmark called MINT (Medical Incremental N-Turn Benchmark) based on five established medical Q\&A datasets including:
1) \textbf{MedQA (N = 600)}~\citep{johri2025evaluation, jin2020diseasedoespatienthave}. We use the version curated in CRAFT-MD, containing 1,804 case vignettes across 12 disease categories. To ensure balanced coverage for each disease, we apply stratified sampling based on disease categories and retain 50 examples per category, resulting in 600 cases; 2) \textbf{MedMCQA (N = 174)}~\citep{pal2022medmcqa}. It has over 194,000 questions from medical entrance examinations across 21 medical subjects including pharmacology, surgery, pediatrics, ophthalmology, dermatology, orthopedics, radiology, psychiatry, and more. We retain diagnosis-focused questions and exclude vignettes shorter than 100 words, resulting in 174 cases; 3) \textbf{Derm-Public (N = 100)}~\citep{johri2025evaluation}. It contains 100 dermatology case vignettes sourced from public educational materials. We manually add diagnostic questions; 4) \textbf{Derm-Private (N = 99)}~\citep{johri2025evaluation}. This dataset has 100 dermatology case vignettes created by three board-certified dermatologists. We remove one mislabeled case and manually add the missing diagnostic question, resulting in 99 cases; and 5) \textbf{MedBullets (N = 62)}~\citep{chen2025medbullets}. It contains 308 clinical questions with expert-written explanations. We select the subset of five-option questions related to medical diagnosis, yielding 62 cases. Additional dataset-level summary statistics are provided in Appendix~\ref{sec:dataset-stats}.

\textbf{Quality Control and Information Preservation.} To minimize information loss and unintended distortions when converting cases from single-turn to multi-turn format, we implement an explicit information-preservation verification protocol for MINT. After decomposing each case into sequential shards, we reconstruct the original input by concatenating all shards into a single prompt (the CONCAT setting) and compare model performance against the original full single-turn input (the FULL setting) using Qwen3-32B. Prediction accuracy remains consistent across the FULL and CONCAT settings (Table \ref{tab:overall-table}), indicating minimal information loss. This confirms that performance differences observed in multi-turn settings reflect changes introduced by the multi-turn structure itself, rather than degradation caused by information loss during conversion.

\textbf{Annotation of Clinical Evidence}. Each shard in the clinically structured setting is annotated using Qwen3-32B with a label drawn from a predefined taxonomy of 13 clinical categories: patient background (e.g., demographics, social history, family history), clinical presentation (e.g., chief complaint, history of present illness, vital signs, physical examination), medical context (e.g., past medical history, medication history), and diagnostic evidence (e.g., laboratory results, imaging, pathology)~\citep{bickley2020bates}. These clinically labeled shards provide a structured and interpretable multi-turn format, grounding evidence sequencing in established clinical categories rather than arbitrary text segmentation.

\textbf{Multi-Turn Construction with Controlled Granularity.} We construct multiple multi-turn variants of each case to support controlled analysis of sequential reasoning: 1) MINT-ClinicalCategoryShard. Using Qwen3-32B with a fixed instruction template (Appendix~\ref{sec:prompting-templates}), we segment each clinical vignette into sequential shards and assign one of 13 clinical category labels to each shard. 2) MINT-FixedTurns. We introduce a turn-granularity setting by repartitioning each case into a fixed number of turns (i.e., 4, 8, 12, 16), enabling fair comparison across turn lengths while preserving the original content and ordering of each case. All variants are generated using Qwen3-32B with a consistent prompting pipeline and deterministic decoding (temperature = 0), ensuring reproducibility across settings. Overall statistics for MINT are reported in Table \ref{tab:data_stats} in Appendix~\ref{sec:dataset-stats}.

\section{Experimental Setup and Results}
\label{sec:exp}

In this section, we evaluate a diverse set of models spanning open-source general-purpose LLMs (Qwen2.5-3B and Qwen3-4B~\citep{yang2025qwen3}, OLMo-2-7B~\citep{olmo2025olmo2}, and GPT-OSS-20B~\citep{openai2025gptoss120bgptoss20bmodel}), commercial frontier LLMs (Claude Sonnet 4.6~\citep{anthropic2026claude46}, GPT-5-mini~\citep{openai2025gpt5}, and o4-mini~\citep{openai2025o4mini}), to medical-finetuned LLMs (Med42-v2-8B, Med42-v2-70B~\citep{med42v2},  MedGemma-1.5-4B, and MedGemma-27B~\citep{sellergren2025medgemma}).

\subsection{Diagnostic Impatience: LLMs Rush to Answer with Insufficient Evidence in Multi-Turn Medical Diagnosis}

\textbf{Setting.} We compare two multi-turn settings that differ only in when the diagnostic question is presented: Ask-Question-First (Q-First) and Ask-Question-Last (Q-Last). In \textbf{Q-First}, the question and answer options are shown upfront, and the model is explicitly instructed to wait until it believes the currently available information is sufficient before answering. As evidence shards are revealed sequentially, the model may continue to hold, answer at any turn, and revise its answer later if subsequent information changes its judgment. In \textbf{Q-Last}, the same shard sequence is shown, but the diagnostic question is withheld until the final turn, so the model answers only once after observing all clinical information. Because both settings use the same cases and shard sequences, this design isolates the effect of answer timing. Here, \emph{hold} denotes choosing not to answer at an intermediate turn, whereas \emph{abstention} denotes failing to produce a final answer after all turns.

\begin{table*}[t]
\centering
\large
\Large
\setlength{\tabcolsep}{2.2pt}
\renewcommand{\arraystretch}{1.2}
\vspace{-3ex}
\resizebox{\textwidth}{!}{%
\begin{tabular}{l ccccccc| ccccccc| ccccccc| ccccccc| ccccccc|}
\toprule
& \multicolumn{7}{c}{\textbf{MedQA}} 
& \multicolumn{7}{c}{\textbf{Derm-Public}}
& \multicolumn{7}{c}{\textbf{Derm-Private}}
& \multicolumn{7}{c}{\textbf{MedBullets}}
& \multicolumn{7}{c}{\textbf{MedMCQA}} \\

& \multicolumn{2}{c}{\textit{Single-turn}} 
& \multicolumn{5}{c|}{\textit{Multi-turn}}
& \multicolumn{2}{c}{\textit{Single-turn}} 
& \multicolumn{5}{c|}{\textit{Multi-turn}}
& \multicolumn{2}{c}{\textit{Single-turn}} 
& \multicolumn{5}{c|}{\textit{Multi-turn}}
& \multicolumn{2}{c}{\textit{Single-turn}} 
& \multicolumn{5}{c|}{\textit{Multi-turn}}
& \multicolumn{2}{c}{\textit{Single-turn}} 
& \multicolumn{5}{c}{\textit{Multi-turn}} \\

& \multicolumn{2}{c}{}
& \multicolumn{3}{c}{\textit{Q-First}} & \multicolumn{2}{c|}{\textit{Q-Last}}
& \multicolumn{2}{c}{}
& \multicolumn{3}{c}{\textit{Q-First}} & \multicolumn{2}{c|}{\textit{Q-Last}}
& \multicolumn{2}{c}{}
& \multicolumn{3}{c}{\textit{Q-First}} & \multicolumn{2}{c|}{\textit{Q-Last}}
& \multicolumn{2}{c}{}
& \multicolumn{3}{c}{\textit{Q-First}} & \multicolumn{2}{c|}{\textit{Q-Last}}
& \multicolumn{2}{c}{}
& \multicolumn{3}{c}{\textit{Q-First}} & \multicolumn{2}{c}{\textit{Q-Last}} \\

\cmidrule(lr){2-8}\cmidrule(lr){9-15}\cmidrule(lr){16-22}
\cmidrule(lr){23-29}\cmidrule(lr){30-36}

\textbf{Model} 
& Full & Conc. & Ini. & Final & \multicolumn{1}{c|}{Abs.} & Ans. & Abs.
& Full & Conc. & Ini. & Final & \multicolumn{1}{c|}{Abs.} & Ans. & Abs.
& Full & Conc. & Ini. & Final & \multicolumn{1}{c|}{Abs.} & Ans. & Abs.
& Full & Conc. & Ini. & Final & \multicolumn{1}{c|}{Abs.} & Ans. & Abs.
& Full & Conc. & Ini. & Final & \multicolumn{1}{c|}{Abs.} & Ans. & Abs. \\
\midrule

\multicolumn{36}{l}{\textbf{\textit{Open-Source General LLMs}}} \\[2pt]

Qwen2.5-3B        
& 53.3 & 51.3 & \cellcolor{red!17}36.2 & \cellcolor{red!24}29.8 & 27.2 & \cellcolor{red!3}50.2 & 0.0
& 64.0 & 69.0 & \cellcolor{red!25}39.1 & \cellcolor{red!36}28.1 & 36.0 & \cellcolor{red!4}60.0 & 0.0
& 74.8 & 75.8 & \cellcolor{red!27}48.0 & \cellcolor{red!45}29.9 & 22.2 & \cellcolor{red!8}66.7 & 0.0
& 40.3 & 41.9 & \cellcolor{red!7}33.3 & \cellcolor{red!14}26.7 & 27.4 & 40.3 & 0.0
& 73.0 & 70.7 & \cellcolor{red!26}47.5 & \cellcolor{red!29}44.5 & 21.3 & \cellcolor{red!5}68.4 & 0.0 \\

Qwen3-4B          
& 63.8 & 65.2 & \cellcolor{red!11}53.0 & \cellcolor{red!5}58.8 & 8.5 & \cellcolor{red!5}58.9 & 0.2
& 82.0 & 79.0 & \cellcolor{red!15}67.0 & \cellcolor{red!8}74.0 & 0.0 & \cellcolor{red!3}79.0 & 0.0
& 87.9 & 84.8 & \cellcolor{red!29}58.8 & \cellcolor{red!15}73.2 & 2.0 & \cellcolor{red!1}86.9 & 0.0
& 48.4 & 48.4 & \cellcolor{red!6}42.9 & \cellcolor{red!4}44.6 & 9.7 & \cellcolor{green!8}56.5 & 0.0
& 84.1 & 80.6 & \cellcolor{red!25}58.7 & \cellcolor{red!12}71.9 & 4.0 & \cellcolor{red!7}77.3 & 0.0 \\

OLMo-2-7B         
& 46.5 & 44.8 & \cellcolor{red!15}31.1 & \cellcolor{red!17}29.4 & 0.3 & \cellcolor{red!8}39.0 & 0.0
& 52.0 & 60.0 & \cellcolor{red!20}32.0 & \cellcolor{red!16}36.0 & 0.0 & \cellcolor{red!6}46.0 & 0.0
& 72.7 & 66.7 & \cellcolor{red!46}27.3 & \cellcolor{red!38}34.3 & 0.0 & \cellcolor{red!19}53.5 & 0.0
& 33.9 & 33.9 & \cellcolor{red!18}16.1 & \cellcolor{red!13}21.0 & 0.0 & \cellcolor{red!8}25.8 & 0.0
& 53.5 & 52.9 & \cellcolor{red!29}24.1 & \cellcolor{red!22}31.6 & 0.0 & \cellcolor{red!4}49.4 & 0.0 \\

GPT-OSS-20B       
& 90.0 & 89.8 & \cellcolor{red!29}61.4 & \cellcolor{red!1}88.8 & 0.2 & \cellcolor{red!4}86.0 & 0.0
& 90.0 & 89.8 & \cellcolor{red!22}68.0 & 90.0 & 0.0 & \cellcolor{red!4}86.0 & 0.0
& 95.0 & 96.0 & \cellcolor{red!30}64.7 & 95.0 & 0.0 & \cellcolor{green!1}95.9 & 0.0
& 70.0 & 67.2 & \cellcolor{red!25}45.2 & \cellcolor{red!1}69.3 & 0.0 & \cellcolor{red!6}64.5 & 0.0
& 94.2 & 93.7 & \cellcolor{red!19}75.7 & 94.2 & 0.6 & \cellcolor{red!1}93.7 & 0.0 \\

\midrule
\multicolumn{36}{l}{\textbf{\textit{Commercial Models}}} \\[2pt]

Claude Sonnet 4.6 
& 94.1 & 93.3 & \cellcolor{red!7}87.7 & \cellcolor{red!1}93.2 & 21.7 & \cellcolor{red!3}91.2 & 2.0
& 94.0 & 96.0 & \cellcolor{red!4}89.8 & \cellcolor{green!4}97.7 & 12.0 & 94.0 & 0.0
& 99.0 & 98.0 & \cellcolor{red!4}94.8 & 99.0 & 2.0 & \cellcolor{green!1}100.0 & 2.0
& 78.7 & 75.4 & \cellcolor{red!7}71.7 & 78.3 & 25.8 & \cellcolor{red!5}73.8 & 1.6
& 97.7 & 97.7 & \cellcolor{red!8}89.6 & \cellcolor{red!2}95.7 & 6.3 & \cellcolor{red!2}95.4 & 1.1 \\

GPT-5-mini        
& 95.5 & 95.0 & \cellcolor{red!19}76.5 & \cellcolor{red!1}94.8 & 0.8 & \cellcolor{red!1}95.0 & 1.5
& 94.0 & 96.0 & \cellcolor{red!16}78.0 & \cellcolor{red!1}93.0 & 0.0 & 94.0 & 0.0
& 98.1 & 97.1 & \cellcolor{red!13}84.8 & \cellcolor{green!1}99.0 & 0.0 & \cellcolor{green!2}100.0 & 0.0
& 80.3 & 77.0 & \cellcolor{red!33}47.5 & \cellcolor{red!2}78.7 & 1.6 & 80.3 & 1.6
& 97.7 & 96.5 & \cellcolor{red!13}84.4 & 97.7 & 0.6 & \cellcolor{red!1}97.1 & 1.7 \\

o4-mini           
& 95.3 & 95.3 & \cellcolor{red!19}76.3 & \cellcolor{red!2}93.6 & 1.3 & \cellcolor{red!1}94.8 & 1.3
& 96.0 & 96.0 & \cellcolor{red!16}80.0 & \cellcolor{red!1}95.0 & 0.0 & \cellcolor{red!1}95.0 & 0.0
& 99.0 & 99.0 & \cellcolor{red!14}84.7 & 99.0 & 1.0 & 99.0 & 0.0
& 77.0 & 80.0 & \cellcolor{red!16}61.0 & \cellcolor{green!3}79.7 & 4.8 & \cellcolor{green!7}83.6 & 1.6
& 98.8 & 97.7 & \cellcolor{red!13}86.1 & \cellcolor{red!2}96.5 & 0.6 & \cellcolor{red!2}97.1 & 0.6 \\

\midrule
\multicolumn{36}{l}{\textbf{\textit{Medical-Domain LLMs}}} \\[2pt]

Med42-v2-8B       
& 72.5 & 70.5 & \cellcolor{red!34}38.7 & \cellcolor{red!24}48.9 & 1.8 & \cellcolor{red!10}62.5 & 0.3
& 84.0 & 76.0 & \cellcolor{red!43}41.0 & \cellcolor{red!21}63.0 & 0.0 & \cellcolor{red!9}75.0 & 0.0
& 85.9 & 89.9 & \cellcolor{red!52}33.7 & \cellcolor{red!14}71.4 & 1.0 & \cellcolor{red!16}69.7 & 0.0
& 61.3 & 51.6 & \cellcolor{red!32}29.0 & \cellcolor{red!15}46.8 & 0.0 & \cellcolor{red!7}54.8 & 0.0
& 89.1 & 86.8 & \cellcolor{red!39}50.0 & \cellcolor{red!18}70.9 & 1.1 & \cellcolor{red!9}79.9 & 0.0 \\

Med42-v2-70B      
& 86.3 & 85.3 & \cellcolor{red!48}38.5 & \cellcolor{red!18}68.3 & 0.0 & \cellcolor{red!7}79.0 & 0.0
& 86.0 & 88.0 & \cellcolor{red!37}49.0 & \cellcolor{red!7}79.0 & 0.0 & 86.0 & 0.0
& 95.0 & 96.0 & \cellcolor{red!66}29.3 & \cellcolor{red!5}89.9 & 0.0 & \cellcolor{red!3}91.9 & 0.0
& 72.6 & 64.5 & \cellcolor{red!48}24.2 & \cellcolor{red!11}61.3 & 0.0 & \cellcolor{red!8}64.5 & 0.0
& 96.5 & 96.0 & \cellcolor{red!46}50.6 & \cellcolor{red!9}87.4 & 0.0 & \cellcolor{red!4}92.5 & 0.0 \\

MedGemma-1.5-4B   
& 67.9 & 68.5 & \cellcolor{red!36}31.8 & \cellcolor{red!36}32.4 & 11.8 & \cellcolor{red!11}57.0 & 1.3
& 69.0 & 71.0 & \cellcolor{red!27}42.4 & \cellcolor{red!27}42.4 & 11.0 & \cellcolor{red!13}56.0 & 0.0
& 81.6 & 83.8 & \cellcolor{red!43}38.4 & \cellcolor{red!44}37.2 & 8.1 & \cellcolor{red!11}70.7 & 0.0
& 56.5 & 51.6 & \cellcolor{red!36}20.8 & \cellcolor{red!38}18.9 & 12.9 & \cellcolor{red!10}46.8 & 0.0
& 82.2 & 82.1 & \cellcolor{red!44}37.8 & \cellcolor{red!42}39.7 & 8.6 & \cellcolor{red!11}71.0 & 2.9 \\

MedGemma-27B      
& 78.3 & 77.8 & \cellcolor{red!5}73.1 & \cellcolor{red!2}76.4 & 50.3 & \cellcolor{red!3}75.6 & 0.3
& 83.0 & 81.0 & \cellcolor{red!4}79.0 & \cellcolor{red!2}80.7 & 38.0 & 83.0 & 0.0
& 91.9 & 91.9 & \cellcolor{red!16}76.2 & \cellcolor{red!12}79.8 & 15.2 & \cellcolor{green!2}93.9 & 0.0
& 67.7 & 66.1 & \cellcolor{red!22}45.8 & \cellcolor{red!22}45.8 & 61.3 & \cellcolor{red!2}66.1 & 0.0
& 89.1 & 89.1 & \cellcolor{red!11}78.5 & \cellcolor{red!9}80.2 & 33.3 & \cellcolor{red!1}87.9 & 0.0 \\

\bottomrule
\end{tabular}
}
\caption{
Performance (\%) under two single-turn settings, \textbf{Full} and \textbf{Conc.}, and two multi-turn settings, \textbf{Ask-Question-First} (\textbf{Q-First}) and \textbf{Ask-Question-Last} (\textbf{Q-Last}). Under \textbf{Q-First}, we report \textbf{Ini.} (accuracy at the first turn where the model answers), \textbf{Final} (accuracy at the final turn after all evidence shards are revealed), and \textbf{Abs.} (abstention rate). Under \textbf{Q-Last}, we report \textbf{Ans.} (final-answer accuracy when the question is revealed only at the last turn). \textbf{Conc.} denotes the concatenated-shards single-turn input. Cell background color indicates the change relative to the same model's \textbf{Full} setting: darker red indicates a larger decrease, and darker green indicates a larger increase.
}
\label{tab:overall-table}
\end{table*}

\textbf{Result.} Across datasets and model families, Table \ref{tab:overall-table} shows a consistent pattern. CONCAT remains close to FULL, indicating minimal information loss from sharding (avg -0.76\%), whereas initial-answer accuracy under Q-First is substantially lower than the single-turn baseline (max 65.7\%). By contrast, Q-Last recovers much of this loss, often approaching FULL (avg -4.1\%). When asking the question first, we observe a large gap (-18.1\% to 60.6\%) between the model's initial and final answer, indicating that initial answering significantly harms performance (up to max 60.6\% drop). When the question is withheld until all clinical information has been presented (Q-Last), performance is consistently higher than Q-First initial-answer accuracy (avg 20.1\%), with models recovering to near single-turn levels. This suggests that models without early answering intent can perform well in multi-turn settings.

\begin{figure}[t]
    \centering
\vspace{-5ex}  
\includegraphics[width=0.95\linewidth]{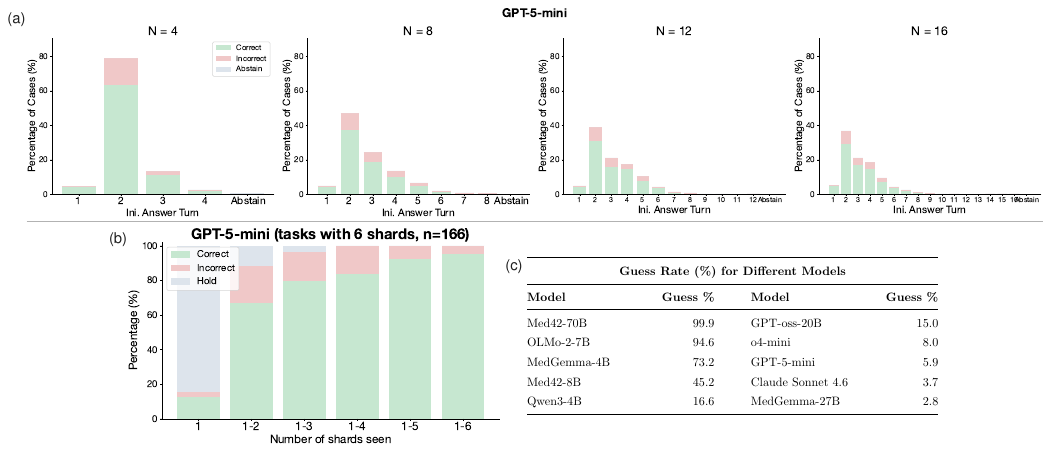}
    \vspace{-0.5em}
    
   \caption{Diagnostic impatience across information accumulation and model types. Colors denote {\fboxsep=1pt\colorbox{bghold}{\vphantom{Ay}hold}}, {\fboxsep=1pt\colorbox{bgcorrect}{\vphantom{Ay}correct}}, and {\fboxsep=1pt\colorbox{bgincorrect}{\vphantom{Ay}incorrect}} responses. (a) Distribution of initial-answer turns under different conversation lengths (N=4, 8, 12, 16) for GPT-5-mini. (b) Turn-level distribution of hold, correct, and incorrect responses for GPT-5-mini by number of shards seen. (c) Guess rates across models.}
    
\vspace{-1em}
    \label{fig:panel_figure}
\end{figure}

We further examine how models behave as information is incrementally revealed. Figure \ref{fig:panel_figure}a shows the distribution of first-answer turns under Q-First across different conversation lengths (N = 4, 8, 12, 16), indicating a persistent pattern of rushing to answer across all settings. In fact, over 55.3\% of answers are given within the first two turns, before sufficient clinical information has been presented. Furthermore, 18.5\% of these initial answers are incorrect, demonstrating that errors are disproportionately concentrated in early turns. Figure~\ref{fig:panel_figure}b illustrates that as GPT-5-mini observes more information shards (1 to 6 out of 6), the proportion of holds decreases while committed answers (correct in green, incorrect in red) increase. This reflects the model's tendency to commit to a diagnosis as evidence accumulates; notably, holding longer can increase accuracy significantly (by 11.7\%).

A striking finding is that some models produce an answer based solely on the diagnostic question, before receiving any clinical information. In the Q-First setting, models are explicitly instructed to WAIT, yet some still answer immediately upon seeing only the diagnostic question and the multiple-choice options. We quantify this behavior as the \textbf{guess rate}, defined as the proportion of cases where the model answers at the initial question-only step. As shown in Figure \ref{fig:panel_figure}c, the guess rate is substantial for several models, reaching 99.9\% for Med42-70B. Conversely, frontier commercial models such as GPT-5-mini (5.9\%) and Claude Sonnet 4.6 (3.7\%) demonstrate a much greater tendency to wait.

\textbf{Summary.} Early answering is the primary driver of performance degradation in multi-turn medical diagnosis, yet deferring the diagnostic question (Q-Last) largely recovers this loss, supporting the view that diagnostic impatience is a major failure mode in our controlled multi-turn setting.

\FloatBarrier

\subsection{Patience Pays: Holding the Diagnosis Restores Accuracy}

\begin{table}[t]
\centering
\vspace{-1em}
\fontsize{6.5}{8}\selectfont
\setlength{\tabcolsep}{1pt}
\renewcommand{\arraystretch}{1.0}
\setlength{\intextsep}{2pt}
\setlength{\columnsep}{6pt}
\begin{tabular}{lccccc}
\toprule
\textbf{Model} & \textbf{Avg Ini. Turn} & \textbf{Abs} & \textbf{Ini.} & \textbf{Full} & \textbf{$\Delta$} \\
\midrule
MedGemma-27B      & $7.2\!\pm\!2.1$ & 46.7 & 77.3 & 86.6 & -9.3  \\
Claude Sonnet 4.6 & $6.5\!\pm\!2.2$ & 22.0 & 90.1 & 96.5 & -6.4  \\
Qwen3-4B          & $4.9\!\pm\!2.5$ & 7.1  & 56.8 & 77.5 & -20.7 \\
o4-mini           & $4.1\!\pm\!1.7$ & 2.2  & 82.0 & 96.1 & -14.1 \\
GPT-5-mini        & $3.7\!\pm\!1.4$ & 1.6  & 80.4 & 96.1 & -15.7 \\
GPT-OSS-20B       & $3.1\!\pm\!1.2$ & 1.1  & 67.8 & 90.0 & -22.2 \\
Med42-8B          & $2.4\!\pm\!2.0$ & 26.4 & 44.8 & 82.8 & -38.0 \\
MedGemma-1.5-4B   & $1.2\!\pm\!0.7$ & 13.7 & 34.4 & 72.0 & -37.6\\
Med42-70B         & $1.0\!\pm\!0.0$ & 0.0  & 40.7 & 90.7 & -50.0 \\
\bottomrule
\end{tabular}
\caption{Comparison of model performance in the 9-turn setting. We report the average turn at which each model produces its first answer ($\mu \pm \sigma$), the abstention rate, the accuracy of the model's initial answer (\textbf{Ini.}), the accuracy when all information is provided (\textbf{Full}), and the accuracy drop from \textbf{Full} to \textbf{Ini.} \textbf{$\Delta$}.}
\label{tab:9turn_condensed}
\end{table}

\begin{figure}[t]
    \centering
\vspace{-0.5em}    \includegraphics[width=0.95\linewidth]{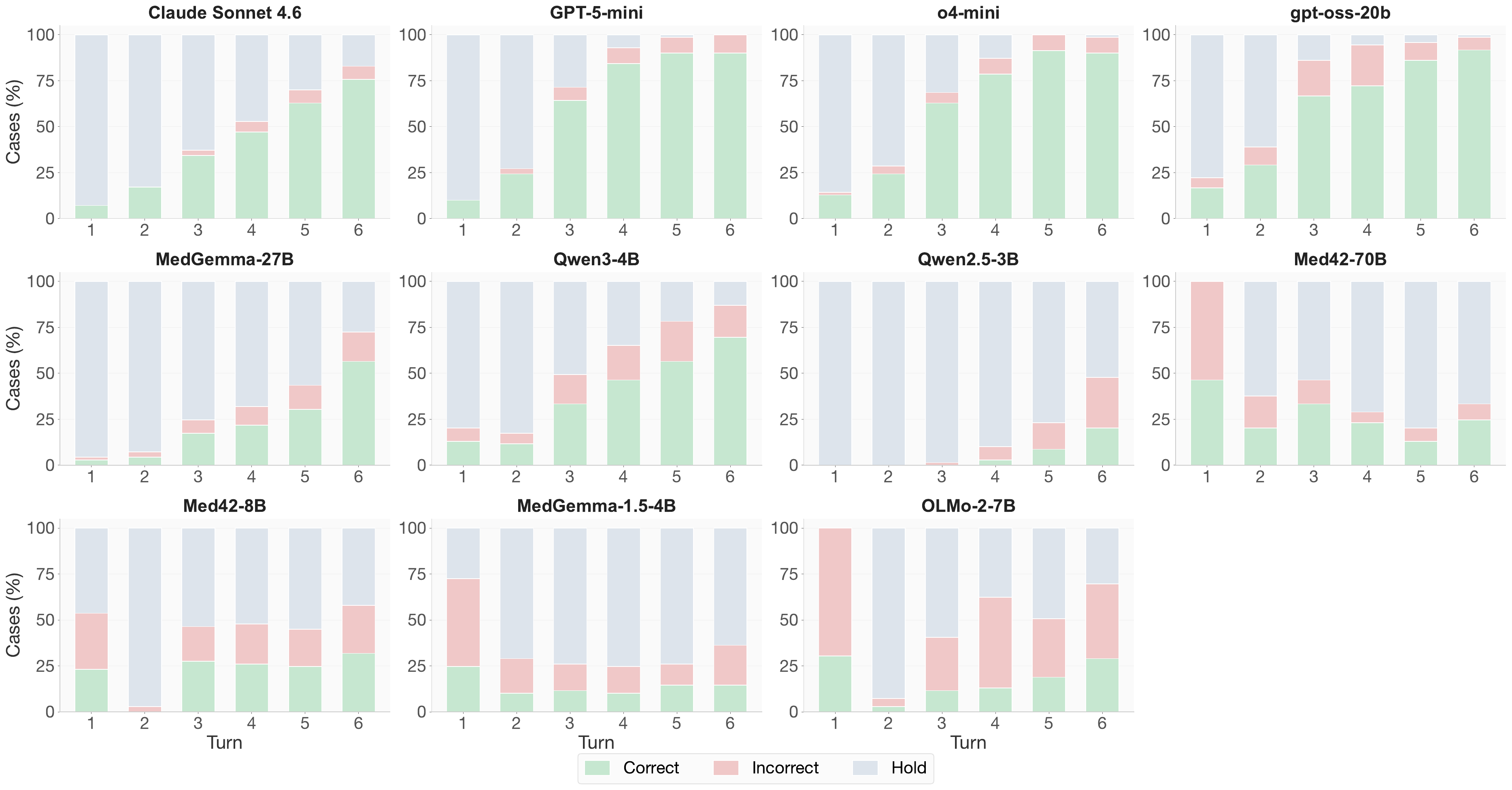}
    \vspace{-0.5em}
    \caption{Proportion of Correct, Incorrect, and Hold responses across six turns for 11 models. For most models, hold responses decrease over turns while the proportion of correct answers increases by the final turn.}
    \vspace{-1em}
    \label{fig:panel_figure2}
\end{figure}

\textbf{Setting.} Motivated by the observation that models frequently answer early and incur large performance drops (avg 24.2\%), we study whether controlling the intent-to-answer can improve performance by comparing Q-First and Q-Last. We examine: 1) the accuracy difference between the model’s first committed answer (Q-First Ini) and Q-Last, 2) answer revisions within a multi-turn conversation using \textbf{F2T} (incorrect-to-correct, measuring the proportion of cases in which the model’s initial committed answer is incorrect but its final committed answer is correct) and \textbf{T2F} (correct-to-incorrect, measuring the proportion of cases in which the model’s initial committed answer is correct but its final committed answer is incorrect), 3) accuracy across different turns (4, 8, 12, 16), 4) model-level statistics such as average Ini  (Avg Ini) and Abstention Rate, 5) specific subsets of cases (e.g., 9-turn cases), and 6) turn-level dynamics in Figure~\ref{fig:panel_figure2}.

\textbf{Result.} Table~\ref{tab:overall-table} shows that Q-Last is consistently higher (avg 20.1\%) than Q-First Ini. across MINT datasets and models, indicating that delayed commitment improves performance. Furthermore, this advantage is also pronounced across the 4-, 8-, 12-, and 16-turn settings, where Q-Last remains consistently higher (avg 14.7\%) than Q-First Ini. in Table~\ref{tab:turn_length_results}. The same pattern is reflected in Table~\ref{tab:9turn_condensed} under the 9-turn setting: models with a later Avg Ini. Turn and a higher abstention rate tend to show smaller drops (9.3\%), whereas models that answer early exhibit much larger drops (50\%) in Full$\rightarrow$Ini. $\Delta$. In short, greater patience is associated with a smaller performance drop.

\begin{table*}[t]
\centering
\small
\setlength{\tabcolsep}{3.2pt}
\renewcommand{\arraystretch}{1.15}
\resizebox{\textwidth}{!}{%
\begin{tabular}{l ccccc| ccccc| ccccc| ccccc| ccccc}
\toprule
& \multicolumn{5}{c|}{\textbf{MedQA}}
& \multicolumn{5}{c|}{\textbf{Derm-Public}}
& \multicolumn{5}{c|}{\textbf{Derm-Private}}
& \multicolumn{5}{c|}{\textbf{MedBullets}}
& \multicolumn{5}{c}{\textbf{MedMCQA}} \\
\cmidrule(lr){2-6}\cmidrule(lr){7-11}\cmidrule(lr){12-16}\cmidrule(lr){17-21}\cmidrule(lr){22-26}
\textbf{\# Turns}
& \textbf{Q.F} & \textbf{FR} & \textbf{T2F} & \textbf{F2T} & \textbf{Q.L}
& \textbf{Q.F} & \textbf{FR} & \textbf{T2F} & \textbf{F2T} & \textbf{Q.L}
& \textbf{Q.F} & \textbf{FR} & \textbf{T2F} & \textbf{F2T} & \textbf{Q.L}
& \textbf{Q.F} & \textbf{FR} & \textbf{T2F} & \textbf{F2T} & \textbf{Q.L}
& \textbf{Q.F} & \textbf{FR} & \textbf{T2F} & \textbf{F2T} & \textbf{Q.L} \\
\cmidrule(lr){2-5}\cmidrule(lr){6-6}\cmidrule(lr){7-10}\cmidrule(lr){11-11}\cmidrule(lr){12-15}\cmidrule(lr){16-16}\cmidrule(lr){17-20}\cmidrule(lr){21-21}\cmidrule(lr){22-25}\cmidrule(lr){26-26}
\midrule
4
& 80.7 & 20.7 & 0.8 & 16.7 & 96.5
& 82.9 & 15.0 & 1.2 & 11.2 & 92.7
& 90.6 & 10.6 & 0.0 & 9.4 & 100.0
& 63.8 & 30.9 & 3.6 & 25.5 & 80.3
& 85.0 & 17.2 & 1.3 & 14.6 & 98.1 \\

8
& 76.3 & 27.0 & 1.5 & 20.3 & 94.9
& 79.3 & 22.0 & 2.4 & 17.1 & 95.1
& 90.1 & 13.6 & 0.0 & 8.6 & 100.0
& 55.9 & 40.7 & 3.4 & 28.8 & 78.3
& 84.1 & 19.2 & 0.6 & 14.7 & 98.1 \\

12
& 77.2 & 28.9 & 1.6 & 19.8 & 93.8
& 83.1 & 16.9 & 1.2 & 10.8 & 92.8
& 84.7 & 17.6 & 0.0 & 15.3 & 100.0
& 60.0 & 34.5 & 1.8 & 20.0 & 76.4
& 86.1 & 20.5 & 2.0 & 12.6 & 98.0 \\

16
& 77.5 & 27.3 & 1.1 & 19.2 & 95.8
& 80.7 & 21.7 & 2.4 & 14.5 & 94.0
& 83.5 & 18.8 & 0.0 & 15.3 & 100.0
& 59.3 & 39.0 & 3.4 & 23.7 & 79.7
& 87.3 & 18.4 & 1.3 & 11.4 & 98.1 \\
\bottomrule
\end{tabular}%
}
\vspace{-0.5em}
\caption{
Performance under different total numbers of turns across five medical benchmarks.
\textbf{Q.F} (Ask Question First) reports accuracy at the model's initial answer, and \textbf{Q.L} (Ask Question Last) reports final diagnostic accuracy when the question is revealed only after all information has been provided.
\textbf{FR} denotes flip rate, \textbf{T2F} denotes true-to-false rate, and \textbf{F2T} denotes false-to-true rate.
}
\label{tab:turn_length_results}
\end{table*}

We further analyze how holding intent affects model performance over multi-turns. Across models, delayed commitment not only reduces premature errors but also preserves the opportunity for later answer revision. This pattern is consistent with the revision dynamics in Table~\ref{table_fr_t2f-f2t}, where \textbf{F2T} is often larger (avg 11.9\%) than \textbf{T2F} in Table~\ref{tab:turn_length_results}, where the same trend holds across different turn settings in ~\ref{tab:turn_length_results} with commercial LLMs having 10.6$\times$ restoration rate (F2T/T2F). Figure~\ref{fig:panel_figure2} provides a turn-level view: for most models, the proportion of correct answers increases by the final turn as hold responses decrease over time. These results show that holding the diagnosis does not merely delay answering, but actively improves reliability by preserving the model's opportunity to self-correct as more evidence is revealed. Additional analyses of the turn-level accuracy plateau and post-plateau gains are provided in Appendix~\ref{app:hold-dynamics}. 

\begin{table*}[t]
\centering
\scriptsize
\setlength{\tabcolsep}{3.0pt}
\renewcommand{\arraystretch}{1.18}
\resizebox{\textwidth}{!}{%
\begin{tabular}{l cccc| cccc| cccc| cccc| cccc}
\toprule
& \multicolumn{4}{c|}{\textbf{MedQA}} 
& \multicolumn{4}{c|}{\textbf{Derm-Public}}
& \multicolumn{4}{c|}{\textbf{Derm-Private}}
& \multicolumn{4}{c|}{\textbf{MedBullets}}
& \multicolumn{4}{c}{\textbf{MedMCQA}} \\
\cmidrule(lr){2-5}\cmidrule(lr){6-9}\cmidrule(lr){10-13}\cmidrule(lr){14-17}\cmidrule(lr){18-21}
\textbf{Model}
& FR & T2F & F2T & RR
& FR & T2F & F2T & RR
& FR & T2F & F2T & RR
& FR & T2F & F2T & RR
& FR & T2F & F2T & RR \\
\midrule
\multicolumn{21}{l}{\textbf{\textit{Open-Source General LLMs}}} \\[2pt]

Qwen2.5-3B
& 52.2 & 16.0 & 9.6 & 0.60
& 53.1 & 20.3 & 9.4 & 0.46
& 50.7 & 22.1 & 3.9 & 0.18
& 46.7 & 17.8 & 11.1 & 0.62
& 45.3 & 14.6 & 11.7 & 0.80 \\

Qwen3-4B
& 27.0 & 4.5 & 10.4 & 2.31
& 13.0 & 0.0 & 7.0 & N/A
& 19.6 & 0.0 & 14.4 & N/A
& 17.9 & 1.8 & 3.6 & 2.00
& 25.8 & 1.8 & 15.0 & 8.33 \\

OLMo-2-7B
& 93.0 & 19.2 & 17.6 & 0.92
& 87.0 & 19.0 & 23.0 & 1.21
& 94.9 & 19.2 & 26.3 & 1.37
& 93.6 & 6.5 & 11.3 & 1.74
& 92.0 & 16.7 & 24.1 & 1.44 \\

GPT-OSS-20B
& 52.3 & 3.5 & 30.9 & 8.83
& 37.0 & 2.0 & 24.0 & 12.00
& 40.4 & 1.0 & 31.3 & 31.30
& 54.8 & 3.2 & 27.4 & 8.56
& 35.8 & 1.2 & 19.7 & 16.42 \\

\midrule
\multicolumn{21}{l}{\textbf{\textit{Commercial Models}}} \\[2pt]

Claude Sonnet 4.6
& 7.7 & 0.6 & 6.2 & 10.33
& 8.0 & 0.0 & 8.0 & N/A
& 4.1 & 0.0 & 4.1 & N/A
& 10.9 & 0.0 & 6.5 & N/A
& 6.8 & 0.0 & 6.1 & N/A \\

GPT-5-mini
& 26.7 & 1.7 & 20.0 & 11.76
& 21.0 & 0.0 & 15.0 & N/A
& 16.2 & 0.0 & 14.1 & N/A
& 42.6 & 0.0 & 31.1 & N/A
& 19.1 & 0.6 & 13.9 & 23.17 \\

o4-mini
& 30.1 & 1.7 & 18.9 & 11.12
& 28.0 & 3.0 & 18.0 & 6.00
& 17.4 & 0.0 & 14.3 & N/A
& 42.4 & 5.1 & 23.7 & 4.65
& 19.1 & 1.7 & 12.1 & 7.12 \\

\midrule
\multicolumn{21}{l}{\textbf{\textit{Medical-Domain LLMs}}} \\[2pt]

Med42-v2-8B
& 54.3 & 9.5 & 19.7 & 2.07
& 50.0 & 4.0 & 26.0 & 6.50
& 56.1 & 1.0 & 38.8 & 38.80
& 61.3 & 9.7 & 27.4 & 2.82
& 48.3 & 5.2 & 26.2 & 5.04 \\

Med42-v2-70B
& 58.3 & 4.3 & 34.2 & 7.95
& 49.0 & 2.0 & 32.0 & 16.00
& 69.7 & 0.0 & 60.6 & N/A
& 75.8 & 1.6 & 38.7 & 24.19
& 55.8 & 2.9 & 39.7 & 13.69 \\

MedGemma-1.5-4B
& 8.4 & 1.6 & 2.1 & 1.31
& 8.2 & 1.2 & 1.2 & 1.00
& 4.7 & 1.2 & 0.0 & 0.00
& 5.7 & 1.9 & 0.0 & 0.00
& 10.9 & 0.6 & 2.6 & 4.33 \\

MedGemma-27B
& 14.8 & 3.0 & 6.4 & 2.13
& 8.1 & 1.6 & 3.2 & 2.00
& 21.4 & 7.1 & 10.7 & 1.51
& 16.7 & 4.2 & 4.2 & 1.00
& 16.4 & 4.3 & 6.0 & 1.40 \\

\bottomrule
\end{tabular}
}
\vspace{-0.5em}
\caption{
Answer-revision behavior across five medical benchmarks.
\textbf{FR} denotes flip rate, \textbf{T2F} denotes true-to-false rate, \textbf{F2T} denotes false-to-true rate, and \textbf{RR} denotes restoration rate, computed as $\text{F2T}/\text{T2F}$, under the multi-turn Q-First setting. When \textbf{T2F} is 0, \textbf{RR} is \textbf{N/A}.
}
\vspace{-1em}
\label{table_fr_t2f-f2t}
\end{table*}

\textbf{Summary.} Holding the diagnosis improves reliability in two ways (Figure~\ref{fig:panel_figure2}): it avoids some early errors and, critically, preserves the opportunity for self-correction to take effect.

\subsection{Strong Trigger: Clinically Salient Information Lures LLMs to Answer Earlier}
\label{sec:lab-result-section}
\textbf{Setting.} To investigate how different types of clinical information influence LLMs’ answering behavior, we conduct a controlled reordering analysis under fixed-turn settings with turn 0 as the questions with the corresponding multiple choices. We hold the underlying cases and laboratory findings constant, and vary only the position of the lab-result shard, placing it early, middle, or late in the multi-turn settings. In the 10-turn setting, the lab-result shard is placed at turn 2 (early), turn 5 (middle), or turn 10 (late). To enable a clinically meaningful analysis, we further group cases into two categories: cases with lab-dependent diseases (e.g., Endocrinology, Infectious Diseases, Hematology and Oncology), where laboratory evidence plays a central role in diagnosis, and cases with non-lab-dependent diseases (e.g., Dermatology, Neurology, Pediatrics), where diagnosis relies less directly on lab results~\citep{caliendo2013better}. All experiments are conducted on MedQA cases that contain laboratory findings (n=249), to avoid confounding from cross-dataset differences. In addition, we analyze answer revision behavior using window size, defined as the turn difference between the first occurrence of two prediction states (e.g., incorrect-to-correct or correct-to-incorrect).

\begin{figure}[t]
    \centering
    \vspace{-2em}\includegraphics[width=1\linewidth]{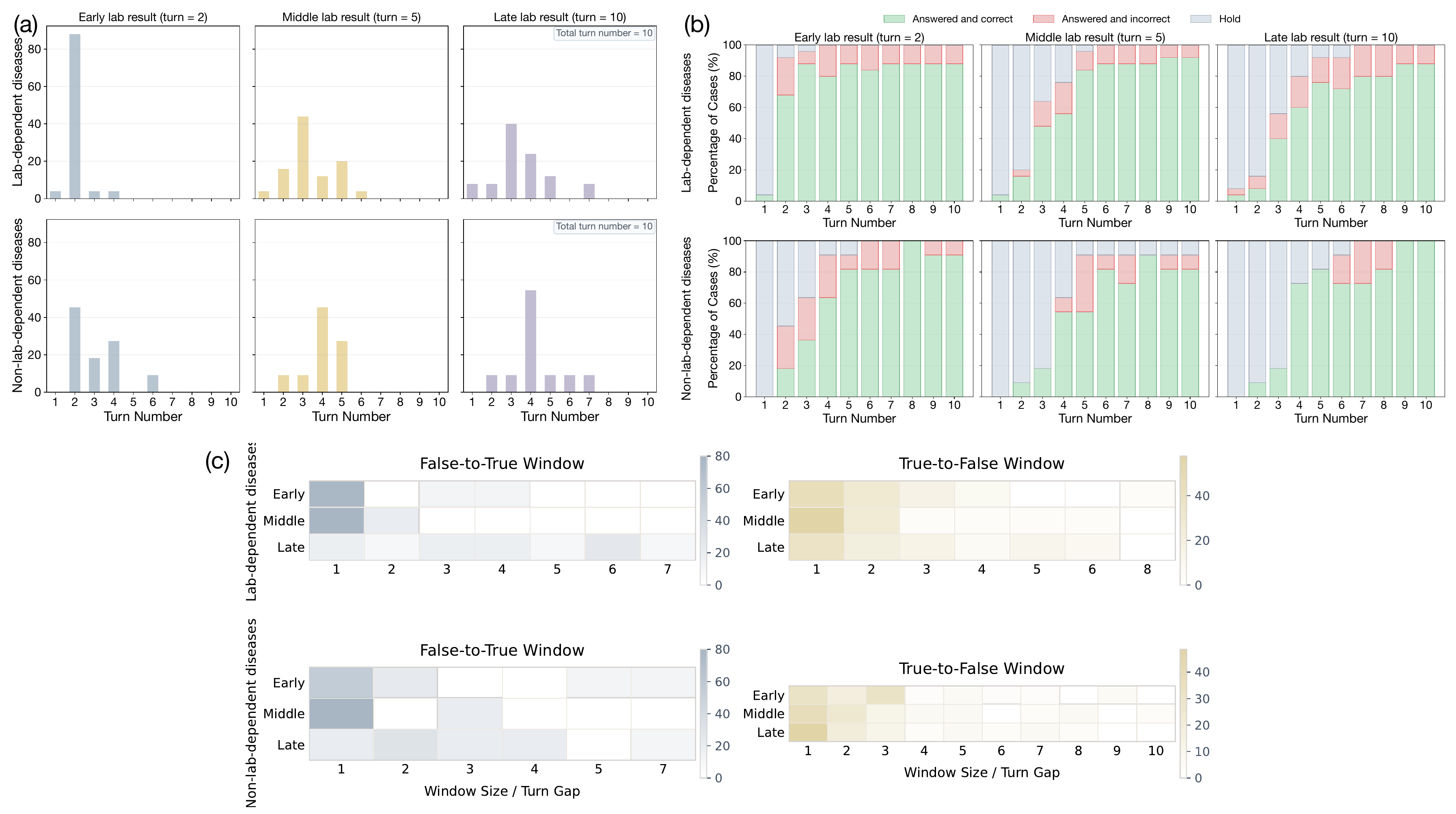}
    \vspace{-1.2em}
    \caption{Effects of lab-result timing on response behavior and self-correction for GPT-5 mini in 10-turn cases. (a) Distribution of the first-answer turn under early, middle, and late lab-result placement, shown separately for lab-dependent and non-lab-dependent diseases. (b) Turn-level response composition for each lab-result order, showing correct, incorrect, and hold states across turns. (c) Heatmaps of False-to-True (left) and True-to-False (right) transitions by window size/turn gap, stratified by disease type and lab-result placement.}
    \vspace{-1em}
    \label{fig:lab-result-panel}
\end{figure}

\textbf{Result.} Prompting alone cannot reliably suppress early answering, and certain clinical information still triggers immediate answers even when models are instructed to wait. In particular, laboratory results act as strong lures that induce early answering. As shown in Figure \ref{fig:lab-result-panel}a, the position of the lab-result shard causes a clear shift in the distribution of initial answer turns. When lab results appear early, models show a strong tendency (75.0\%) to answer immediately, whereas moving the same lab information to later turns substantially diminishes this urge (13.9\% vs. 8.3\%), compared with middle and late placements. This effect is especially pronounced for lab-dependent diseases (Figure \ref{fig:lab-result-panel}a): when lab results appear at turn 2, 88.0\% of lab-dependent cases produce their first answer at that turn, compared with 45.5\% for non-lab-dependent cases. When the lab-result shard is moved to the middle (turn 5), immediate answering weakens significantly (16.0\%), and in the late condition (turn 10), only 8.0\% of lab-dependent cases answer immediately after the lab-result turn. Importantly, this behavior is not limited to lab-driven diagnoses: even for non-lab-dependent diseases, placing lab results early shifts answering to earlier turns (45.5\% vs. 9.1\% in middle and late settings), indicating that laboratory evidence broadly acts as a strong trigger for premature answering. Further descriptive statistics and additional experimental details for this analysis are reported in Appendix~\ref{sec:lab-test}.

Intent-to-answer driven by strong lures directly affects diagnostic outcomes. Figure~\ref{fig:lab-result-panel}b shows that when lab results appear early, the proportion of unanswered cases drops sharply immediately after the lab-result turn, indicating that many cases transition into committed answers at that point. However, this earlier commitment is not uniformly beneficial: part of this shift corresponds to incorrect answers, showing that lab-induced early answering can accelerate both correct and incorrect decisions. When lab results are moved to later positions, answering becomes more gradual and early-answer errors are reduced. Notably, the impact differs across disease types. For lab-dependent diseases, early answering induced by lab results leads to relatively smaller performance degradation, as lab information is often diagnostically central. In contrast, for non-lab-dependent diseases, premature answering triggered by lab results leads to substantially higher error rates, since diagnosis relies less directly on laboratory evidence.
Finally, we analyze how lab-result position affects answer revision dynamics (Figure~\ref{fig:lab-result-panel}c). Answer changes are mostly concentrated within small window sizes, indicating that revisions are typically local. When lab results appear early or in the middle, F2T transitions are highly concentrated at window size 1 (e.g., 77.8\% and 80.0\% for lab-dependent diseases), suggesting immediate self-correction following the introduction of lab evidence. In contrast, when lab results appear late, self-correction becomes more distributed across larger window sizes, indicating that corrections unfold more gradually over subsequent turns.

\textbf{Summary.} These results show that answering behavior depends not only on internal answer tendency, but also on the type of clinical evidence. Clinically salient signals such as laboratory results can act as strong triggers that induce early answering and reshape both error patterns and self-correction dynamics in multi-turn diagnosis. 

Representative qualitative case studies of these behaviors, including rush-to-answer, self-correction, and lab-triggered premature answering, are provided in Appendix~\ref{sec:CaseStudies}.

\section{Conclusion}
In this work, we build MINT, a controlled benchmark for evaluating LLM behaviors in multi-turn medical diagnosis.
Through three complementary lenses, Hold, Self-Correction, and Lure, we demonstrate that multi-turn performance degradation is driven primarily by premature commitment rather than insufficient diagnostic capability. 
Deferring the diagnostic question recovers accuracy to near single-turn levels, incorrect-to-correct revisions dominate the reverse by up to 10.6$\times$, and early placement of laboratory results reliably triggers the premature answering that causes diagnostic errors. 
These findings reframe the central challenge: the bottleneck is not whether a model can reason, but whether it can govern commitment timing as evidence unfolds. 
An important direction for future work is to develop commitment-aware diagnostic systems that explicitly optimize \emph{when} to answer alongside \emph{what} to diagnose.


\newpage
\section*{Reproducibility Statement }
All five datasets used to construct MINT (MedQA, MedMCQA, Derm-Public, Derm-Private, and MedBullets) are publicly available. Detailed descriptions of each dataset, including preprocessing steps, annotation procedures, and shard construction pipelines, are provided in Appendix~\ref{sec:dataset-stats} and~\ref{sec:prompting-templates}. Section~\ref{sec:exp} describes the full experimental setup, including model configurations, prompting templates, and decoding parameters. All multi-turn variants are generated using deterministic decoding (temperature = 0) to ensure reproducibility across settings. The complete MINT benchmark, including all shard variants, evaluation scripts, and prompting templates, will be open-sourced upon publication.

\section*{Ethics Statement}
MINT is constructed from publicly available medical question-answering datasets and does not involve patient-identifiable data or human subjects. Our evaluation involves the use of commercial LLM APIs, including OpenAI and Anthropic services. We adhere to the data usage guidelines of each respective API provider. This work is intended to improve understanding of LLM behavior in medical diagnosis settings and does not constitute clinical validation or endorsement of any model for clinical deployment. We do not foresee additional ethical concerns arising from this study.

\bibliography{colm2026_conference}
\bibliographystyle{colm2026_conference}

\newpage
\appendix
\section{Limitations}
Several limitations remain. First, our benchmark is constructed by converting existing single-turn medical QA cases into sequential evidence streams, rather than collecting naturally occurring doctor–patient interactions. Second, our central intervention, holding commitment by revealing the diagnostic question only after all evidence has been presented, is primarily an analytic control, not a realistic deployment protocol. Third, although we explicitly verify information preservation by comparing FULL and CONCAT, this verification is still limited. Fourth, the benchmark remains limited in coverage and ecological diversity. Fifth, our evaluation focuses on multiple-choice diagnostic commitment, which is analytically clean but narrower than real clinical reasoning. Sixth, while our results strongly support the importance of premature commitment, they do not fully eliminate other explanations for multi-turn degradation. Our controlled setup isolates answer timing, turn granularity, and evidence type, but real multi-turn failure may also reflect longer-horizon state tracking, memory errors, prompt sensitivity, instruction drift, or interaction effects between reasoning style and dialogue context. All these limitations point out interesting future work directions, especially the bold direction of uncovering mechanism of whether LLMs can decide when to wait, when to answer, and when to revise under partial information. In this sense, reliable multi-turn medical diagnosis may require not only stronger reasoning models, but more decision-aware and action-oriented systems.

\section{Dataset Details}
\label{sec:dataset-stats}
Table~\ref{tab:data_stats} provides additional summary statistics for the five source datasets included in MINT. We report the number of cases, the average and median number of turns after sharding, the average case length, and the number and proportion of laboratory-result shards.

\begin{table}[!htbp]
\centering
\scriptsize
\setlength{\tabcolsep}{3pt}
\renewcommand{\arraystretch}{1.1}
\begin{tabular}{lccccc}
\toprule
\textbf{MINT} & \textbf{N} & \textbf{Mean/Turn} & \textbf{Median/Turn} & \textbf{Len./Case} & \textbf{Lab/Shard} \\
\midrule
MedQA~\citep{johri2025evaluation}        & 600  & 9.92  & 10 & 128.39 & 249 (41.5\%) \\
MedMCQA~\citep{pal2022medmcqa}        & 174  & 8.75  & 8  & 88.98  & 86 (49.4\%) \\
Derm-Public~\citep{johri2025evaluation}  & 100  & 7.79  & 8  & 155.29 & 8 (8.0\%) \\
Derm-Private~\citep{johri2025evaluation} & 99   & 7.75  & 8  & 139.31 & 7 (7.1\%) \\
MedBullets~\citep{chen2025medbullets} & 62   & 10.74 & 11 & 169.63 & 20 (32.3\%) \\
\midrule
Overall                               & 1035 & 9.36  & 9  & 124.63 & 370 (35.7\%) \\
\bottomrule
\end{tabular}
\caption{Descriptive statistics of the source dataset.}
\label{tab:data_stats}
\end{table}

\section{Additional Turn-Level Dynamics of Holding and Self-Correction}
\label{app:hold-dynamics}
\begin{figure}[ht!]
    \centering
    \includegraphics[width=0.95\linewidth]{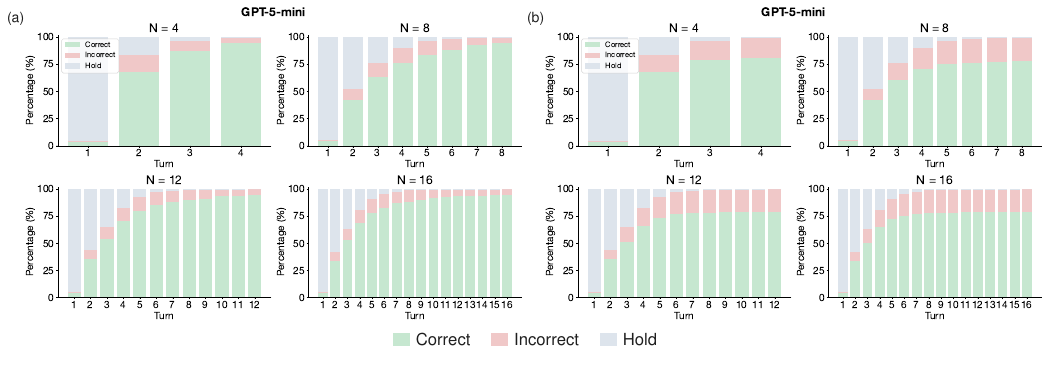}
    \caption{(a) Running answer states across turns for GPT-5-mini under different shard counts ($N = 4, 8, 12, 16$). As more evidence is revealed, the model’s current answer state can still improve, with some initially incorrect states later corrected. (b) Cumulative first-answer outcomes across turns. Most first committed answers occur early, and many of these early commitments are incorrect.}
    \label{fig:panel_figure2_bc}
\end{figure}

Figure~\ref{fig:panel_figure2_bc}a shows that initial-answer accuracy tends to plateau after the first few turns, suggesting that the marginal gain in first committed answers becomes limited once the model has already formed an early diagnosis. However, Figure~\ref{fig:panel_figure2_bc}b shows that overall accuracy can continue to improve beyond this plateau when later answer revisions are allowed. This divergence arises because false-to-true transitions occur more often than true-to-false transitions, so additional evidence more often corrects an earlier mistake than overturns a correct early answer. In this sense, post-plateau gains are largely driven by self-correction rather than by continued improvement in first-answer accuracy.

\section{Additional Analysis of Lab Test Ordering Patterns}
\label{sec:lab-test}
\subsection{Distribution Analysis}
\begin{figure}[ht!]
    \centering
    \includegraphics[width=0.95\linewidth]{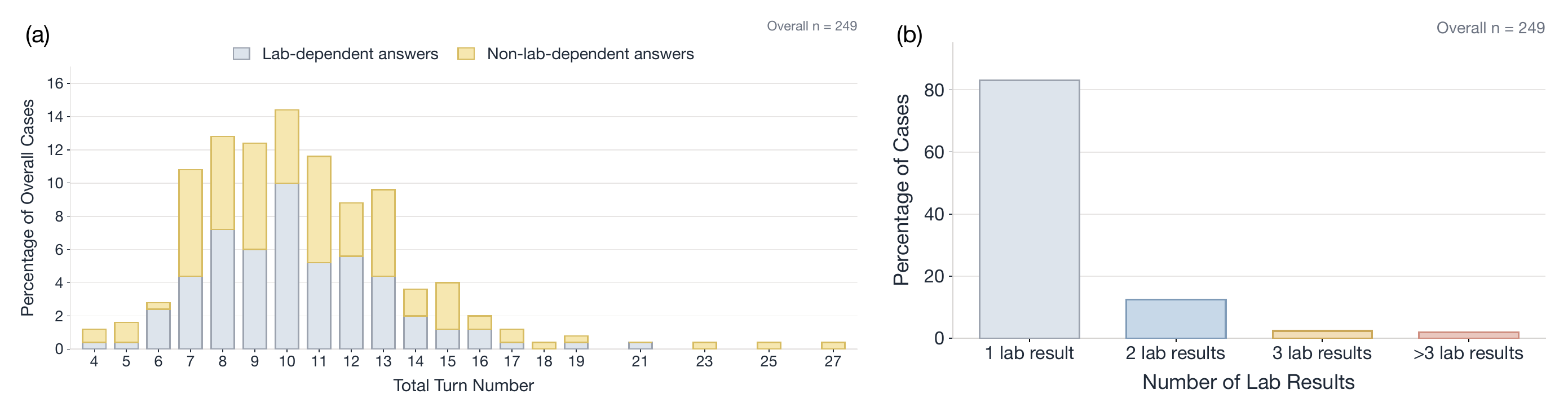}
    \caption{Distribution of lab-result cases by total turn count and by number of lab results.}
    \label{fig:lab-result-desc-bucket-panel}
\end{figure}

Figure~\ref{fig:lab-result-desc-bucket-panel} shows the distribution of lab-result cases. Approximately 80\% of cases include a single shard containing a lab result record. Lab-result cases are most concentrated between 7 and 12 turns.

\subsection{Disease-Level Results by Lab-Result Order}
To provide a more fine-grained view of the lab-result reordering analysis, Table~\ref{tab:lab-result-per-12-diseases} reports first-answer accuracy under the Ask Question First setting, broken down by disease and lab-result order. Results are shown for early, middle, and late lab-result placement with the number of turns fixed at 10. Bold disease names denote lab-dependent diseases.
\begin{table*}[ht]
\centering
\scriptsize
\setlength{\tabcolsep}{4pt}
\renewcommand{\arraystretch}{1.15}
\resizebox{\textwidth}{!}{%
\begin{tabular}{l ccc| ccc| ccc| ccc}
\toprule
\textbf{Disease} & \multicolumn{3}{c|}{\textbf{GPT-5-mini}} & \multicolumn{3}{c|}{\textbf{Claude Sonnet 4.6}} & \multicolumn{3}{c|}{\textbf{Qwen3-4B}} & \multicolumn{3}{c}{\textbf{MedGemma-27B}} \\
& \textbf{Early} & \textbf{Middle} & \textbf{Late} & \textbf{Early} & \textbf{Middle} & \textbf{Late} & \textbf{Early} & \textbf{Middle} & \textbf{Late} & \textbf{Early} & \textbf{Middle} & \textbf{Late} \\
\midrule
Cardiovascular System & 63.64 & \cellcolor{green!27}90.91 & \cellcolor{green!27}90.91 & 45.45 & \cellcolor{green!18}63.64 & \cellcolor{green!18}63.64 & 54.55 & \cellcolor{green!9}63.64 & \cellcolor{green!9}63.64 & 27.27 & \cellcolor{green!9}36.36 & 27.27 \\
Dermatology & 54.55 & \cellcolor{green!9}63.64 & \cellcolor{green!36}90.91 & 54.55 & \cellcolor{red!9}45.45 & \cellcolor{green!9}63.64 & 18.18 & \cellcolor{green!9}27.27 & \cellcolor{green!9}27.27 & 45.45 & \cellcolor{green!9}54.55 & \cellcolor{red!18}27.27 \\
\textbf{Endocrinology} & 86.11 & \cellcolor{green!3}88.89 & \cellcolor{red!6}80.56 & 72.22 & \cellcolor{green!22}94.44 & \cellcolor{green!3}75.00 & 36.11 & \cellcolor{green!6}41.67 & \cellcolor{green!8}44.44 & 36.11 & \cellcolor{red!3}33.33 & \cellcolor{green!3}38.89 \\
Gastrointestinal System & 79.17 & \cellcolor{red!4}75.00 & \cellcolor{green!4}83.33 & 75.00 & \cellcolor{green!8}83.33 & \cellcolor{red!17}58.33 & 54.17 & \cellcolor{red!8}45.83 & 54.17 & 45.83 & \cellcolor{green!13}58.33 & \cellcolor{green!8}54.17 \\
\textbf{Hematology and Oncology} & 87.88 & \cellcolor{red!6}81.82 & 87.88 & 81.82 & \cellcolor{green!3}84.85 & \cellcolor{red!9}72.73 & 60.61 & \cellcolor{red!3}57.58 & \cellcolor{red!12}48.48 & 42.42 & \cellcolor{green!6}48.48 & \cellcolor{red!6}36.36 \\
\textbf{Infectious Diseases} & 57.89 & \cellcolor{green!11}68.42 & \cellcolor{green!11}68.42 & 63.16 & \cellcolor{red!5}57.89 & \cellcolor{green!5}68.42 & 26.32 & \cellcolor{green!16}42.11 & \cellcolor{green!5}31.58 & 36.84 & 36.84 & \cellcolor{red!21}15.79 \\
Neurology & 60.00 & \cellcolor{green!10}70.00 & \cellcolor{green!20}80.00 & 50.00 & \cellcolor{green!30}80.00 & \cellcolor{green!40}90.00 & 30.00 & 30.00 & 30.00 & 20.00 & \cellcolor{green!20}40.00 & \cellcolor{green!20}40.00 \\
Obstetrics and Gynecology & 65.22 & \cellcolor{red!9}56.52 & \cellcolor{red!17}47.83 & 69.57 & \cellcolor{red!4}65.22 & \cellcolor{red!9}60.87 & 39.13 & \cellcolor{green!4}43.48 & \cellcolor{green!4}43.48 & 43.48 & \cellcolor{green!17}60.87 & \cellcolor{red!4}39.13 \\
Other & 83.33 & \cellcolor{red!33}50.00 & 83.33 & 83.33 & \cellcolor{red!17}66.67 & 83.33 & 83.33 & 83.33 & 83.33 & 66.67 & \cellcolor{red!17}50.00 & \cellcolor{red!17}50.00 \\
Pediatrics and Neonatology & 80.00 & \cellcolor{green!10}90.00 & 80.00 & 60.00 & \cellcolor{green!10}70.00 & \cellcolor{red!20}40.00 & 60.00 & \cellcolor{red!10}50.00 & 60.00 & 40.00 & \cellcolor{green!20}60.00 & 40.00 \\
Rheumatology & 84.00 & \cellcolor{red!4}80.00 & \cellcolor{red!8}76.00 & 80.00 & \cellcolor{green!4}84.00 & 80.00 & 60.00 & \cellcolor{red!4}56.00 & \cellcolor{red!4}56.00 & 36.00 & \cellcolor{green!16}52.00 & \cellcolor{green!16}52.00 \\
\textbf{Urology and Nephrology} & 82.93 & \cellcolor{green!2}85.37 & \cellcolor{red!10}73.17 & 75.61 & \cellcolor{red!10}65.85 & \cellcolor{red!10}65.85 & 56.10 & \cellcolor{green!2}58.54 & \cellcolor{green!7}63.41 & 36.59 & \cellcolor{red!5}31.71 & 36.59 \\
\bottomrule
\end{tabular}
}
\caption{First-answer accuracy (\%) under \textbf{Ask Question First}, broken down by disease and lab-result order. Results are shown for \textbf{Early}, \textbf{Middle}, and \textbf{Late} lab-result placement, with the number of turns fixed at 10. Bold disease names denote lab-dependent diseases.}
\label{tab:lab-result-per-12-diseases}
\end{table*}

\subsection{Overall Results by Lab-Result Order}
Table~\ref{tab:model_order_performance} summarizes overall model performance under different lab-result orders. We report results for both Ask Question First and Ask Question Last. For Ask Question First, we report initial-answer accuracy, final-answer accuracy, and abstention rate; for Ask Question Last, we report final-answer accuracy and abstention rate. We found that when the question is asked late, reserving clinically salient information for later turns helps prevent a catastrophic accuracy drop (up to 23.3\%).

\begin{table*}[t]
\centering
\scriptsize
\setlength{\tabcolsep}{6pt}
\renewcommand{\arraystretch}{1.12}
\begin{tabular}{llccccc}
\toprule
\textbf{Model} & \textbf{Order} & \multicolumn{3}{c}{\textbf{Ask Q First}} & \multicolumn{2}{c}{\textbf{Ask Q Last}} \\
\cmidrule(lr){3-5}\cmidrule(lr){6-7}
 & & \textbf{Ini.} & \textbf{Final} & \textbf{Abs.} & \textbf{Final} & \textbf{Abs.} \\
\midrule

\multirow{3}{*}{GPT-5-mini}
& Early  & 76.83 & 95.12 & 0.00 & 96.34 & 0.00 \\
& Middle & \cellcolor{green!3}77.64 & \cellcolor{red!3}94.31 & \cellcolor{green!2}0.40 & \cellcolor{red!2}95.93 & \cellcolor{green!2}0.40 \\
& Late   & \cellcolor{green!1}77.11 & \cellcolor{green!7}96.79 & \cellcolor{green!2}0.40 & \cellcolor{red!3}95.58 & \cellcolor{green!2}0.40 \\
\midrule

\multirow{3}{*}{Claude Sonnet 4.6}
& Early  & 71.54 & 80.89 & 7.23 & 93.50 & 0.40 \\
& Middle & \cellcolor{green!16}75.61 & \cellcolor{red!10}78.46 & \cellcolor{green!18}11.65 & \cellcolor{red!13}90.24 & \cellcolor{red!2}0.00 \\
& Late   & \cellcolor{red!10}69.08 & \cellcolor{red!39}71.08 & \cellcolor{green!22}12.85 & \cellcolor{red!5}92.37 & \cellcolor{green!3}1.20 \\
\midrule

\multirow{3}{*}{Qwen3-4B}
& Early  & 48.19 & 58.23 & 1.20 & 59.04 & 0.40 \\
& Middle & \cellcolor{green!6}49.80 & \cellcolor{red!8}56.22 & \cellcolor{green!13}4.42 & \cellcolor{green!2}59.44 & 0.40 \\
& Late   & \cellcolor{green!8}50.20 & \cellcolor{red!19}53.41 & \cellcolor{green!10}3.61 & \cellcolor{red!2}58.63 & 0.40 \\
\midrule

\multirow{3}{*}{MedGemma-27B}
& Early  & 58.23 & 64.66 & 8.43 & 77.11 & 3.21 \\
& Middle & \cellcolor{green!14}61.85 & \cellcolor{green!8}66.67 & \cellcolor{green!6}10.04 & \cellcolor{green!2}77.51 & \cellcolor{green!6}4.82 \\
& Late   & \cellcolor{green!3}59.04 & 64.66 & \cellcolor{green!21}13.65 & \cellcolor{green!11}79.92 & \cellcolor{green!2}3.61 \\

\bottomrule
\end{tabular}
\caption{Overall performance (\%) under different lab-result orders. Results are shown for \textbf{Early}, \textbf{Middle}, and \textbf{Late} lab-result placement. We report initial-answer accuracy, final-answer accuracy, and abstention rate for \textbf{Ask Question First}, and final-answer accuracy and abstention rate for \textbf{Ask Question Last}.}
\label{tab:model_order_performance}
\end{table*}

\subsection{Representative Trajectories Under Lab-Result Reordering}
\begin{figure}[ht!]
    \centering
    \includegraphics[width=0.85\linewidth]{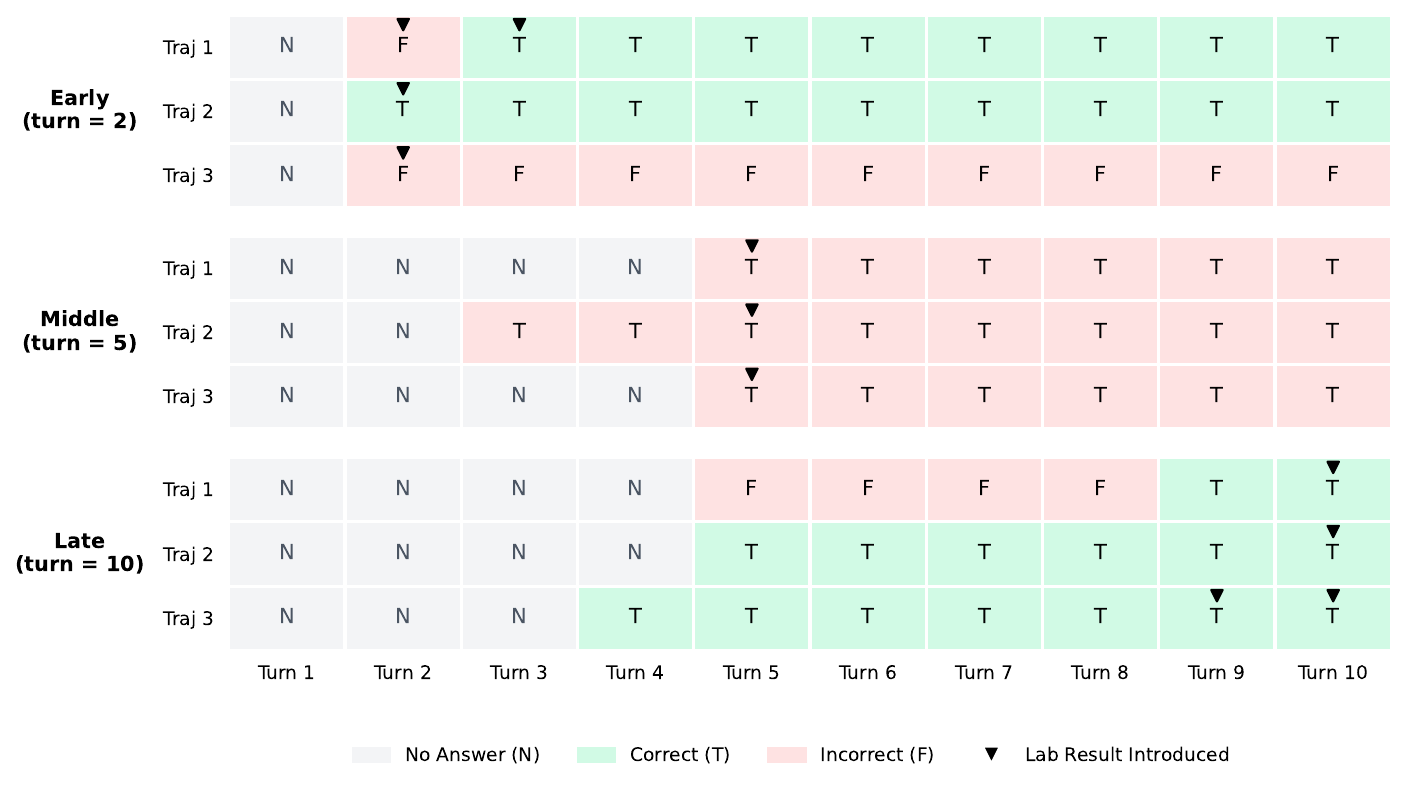}
    \caption{Representative response trajectories under different lab-result orders. For GPT-5-mini on 10-turn cases, each row shows an example case trajectory when the key lab result is introduced early, middle, or late. Cells indicate whether the model gives no answer (N), a correct answer (T), or an incorrect answer (F) at each turn; outlined cells mark the turn where the lab result is revealed.}
    \label{fig:lab-result-example}
\end{figure}

We provide representative response trajectories to illustrate how lab-result timing changes answer behavior at the turn level. Figure~\ref{fig:lab-result-example} shows three example trajectories under early-, middle-, and late lab-result. These examples visually complement the aggregate results in Section~\ref{sec:lab-result-section} by showing how early lab disclosure can trigger immediate commitment, while later disclosure can delay answering and sometimes preserve opportunities for correction.

\newpage

\section{Case Studies}
\label{sec:CaseStudies}
\begin{figure}[ht!]
    \centering
    \includegraphics[width=0.95\linewidth]{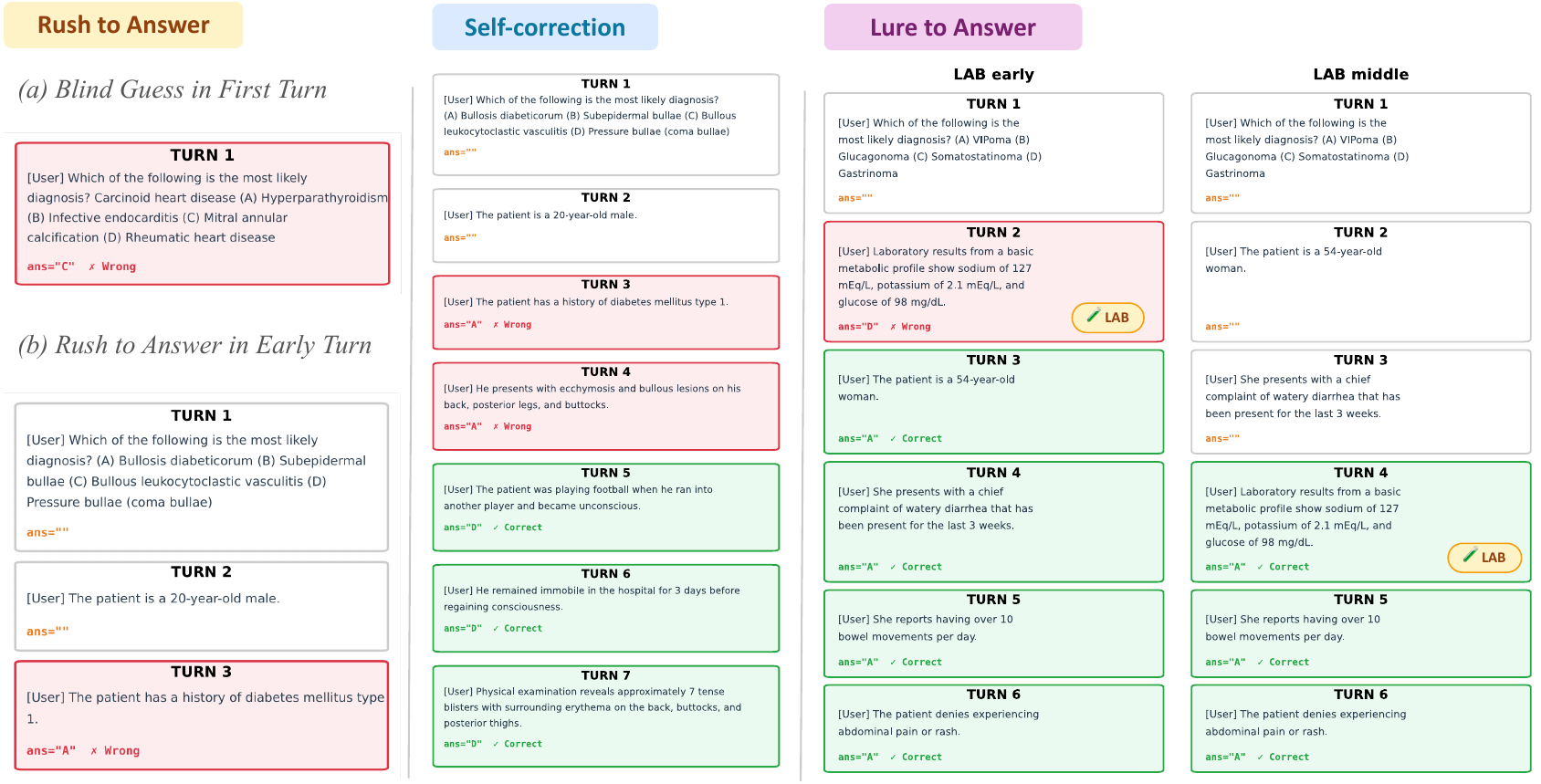}
    \caption{Representative behavioral patterns observed in Claude Sonnet~4.6. \textbf{Left:} \textit{Rush to Answer}. (a)~The model guesses at Turn~1 with no clinical context, and (b)~it answers at an early turn with minimal information. \textbf{Middle:} \textit{Self-correction}. The model initially answers incorrectly but revises to the correct option over subsequent turns. \textbf{Right:} \textit{Lure to Answer}. The same question under two lab result placements (early vs.\ middle, marked with \textsc{Lab} badges); early placement triggers earlier responses. Green and red borders indicate correct and incorrect answers, respectively.}
    \label{fig:case-studies}
\end{figure}

\textbf{Rush to Answer.}
Figure~\ref{fig:case-studies} (left) illustrates models may rush to answer without waiting for sufficient information. In case~(a), the model answers at Turn~1 before any clinical information is revealed, guessing based on the question alone. In case~(b), the model waits through the first turn but answers at Turn~3 with only basic demographic and history information. Both cases suggest that models may rush to answer without waiting for sufficient information.

\textbf{Self-correction.}
The middle panel of Figure~\ref{fig:case-studies} shows the model's capacity for self-correction. The model initially provides a wrong answer at Turn~3 and persists through Turn~4, but revises it to the correct option by Turn~5 as more information are revealed. This suggests that premature responses are not always final. Models may recover when subsequent evidence is provided.

\textbf{Lab Results as Answering Triggers.}
The right panel of Figure~\ref{fig:case-studies} compares two orderings of the same question that differ only in when laboratory results are revealed (marked with \textsc{Lab} badges). When lab results appear at an early turn (left), the model answers as early as Turn~2; when the same laboratory results are deferred to a middle turn (right), the model waits longer and answers correctly on the first attempt. This indicates that laboratory findings serve as strong answering signals that can lure models into responding earlier, regardless of whether the surrounding clinical context is sufficient.

\section{Prompting Templates}
\label{sec:prompting-templates}

\subsection{Sharding Prompt}
\begin{promptbox}[Sharding Prompt]
\ttfamily\small

You are helping to convert a medical multiple-choice question into a multi-turn conversational format for a reasoning dataset.

Your task is to:
\begin{enumerate}
\item Segment the question into clinically meaningful pieces
\item Assign each segment a clinical category label
\item Rephrase each segment into a natural conversational shard
\end{enumerate}

\textcolor{red}{\textbf{Clinical categories:}}
\begin{itemize}
\item DEMOGRAPHICS: patient age, sex, occupation
\item CHIEF\_COMPLAINT: primary symptom or reason for visit, duration
\item HISTORY\_OF\_PRESENT\_ILLNESS: onset, progression, associated symptoms
\item PAST\_MEDICAL\_HISTORY: prior diagnoses, previous episodes
\item MEDICATION\_HISTORY: current or recent medications
\item SOCIAL\_HISTORY: smoking, alcohol, drug use, living situation
\item FAMILY\_HISTORY: relevant family conditions
\item VITAL\_SIGNS: temperature, blood pressure, pulse, respiration, oxygen saturation
\item PHYSICAL\_EXAM: findings on inspection, palpation, auscultation, percussion
\item LAB\_RESULTS: blood tests, urine tests, cultures
\item IMAGING\_OR\_DIAGNOSTICS: X-ray, CT, MRI, EKG, ultrasound
\item PATHOLOGY\_OR\_SMEAR: biopsy, smear, histology findings
\item OTHER\_CLINICAL\_INFO: any clinical information that does not fit the above categories
\item QUESTION\_STEM: the final diagnostic question sentence
\end{itemize}

\textcolor{red}{\textbf{Segmentation guidelines:}}
\begin{itemize}
\item Each segment must belong to exactly ONE clinical category.
\item Each segment must be copied VERBATIM from the original question. Do not rephrase or add words.
\item Segments must be non-overlapping and together cover the entire question text including the final question sentence.
\item If multiple findings belong to the same clinical category and appear in the same sentence or clause, keep them as ONE segment rather than splitting.
\item If one sentence contains information from multiple categories, split it at the category boundary.
\item Do not start a segment with conjunctions like ``and'' or ``but'' --- merge them with the previous segment instead.
\item Assign a sequential id to each segment starting from 1, reflecting the original order in the question.
\item The answer choices are NOT segmented --- ignore them.
\end{itemize}

\textcolor{red}{\textbf{Rephrasing guidelines:}}
\begin{itemize}
\item Rephrase each segment into a complete, natural clinical sentence.
\item Preserve the original order of segments --- do not reorder.
\item Do not reveal or hint at the correct answer anywhere in the shards.
\item Each shard must have proper capitalization and punctuation.
\end{itemize}

\textcolor{red}{\textbf{Output format:}}
\begin{verbatim}
{
  "shards": [
    {
      "id": 1,
      "segment": "[verbatim excerpt from the question]",
      "shard": "conversational rephrasing of this segment",
      "label": "CATEGORY"
    }
  ]
}
\end{verbatim}

Example:

Question: ``A 66-year-old man presents with fatigue. He has hypertension and drinks 3 beers a day. His temperature is 37.1°C and pulse is 78. Which of the following is the most likely diagnosis?''

Output:
\begin{verbatim}
{
  "shards": [
    {
      "id": 1,
      "segment": "A 66-year-old man",
      "shard": "The patient is a 66-year-old man.",
      "label": "DEMOGRAPHICS"
    },
    {
      "id": 2,
      "segment": "presents with fatigue.",
      "shard": "He presents with a chief complaint of fatigue.",
      "label": "CHIEF_COMPLAINT"
    },
    {
      "id": 3,
      "segment": "He has hypertension",
      "shard": "The patient has a history of hypertension.",
      "label": "PAST_MEDICAL_HISTORY"
    },
    {
      "id": 4,
      "segment": "and drinks 3 beers a day.",
      "shard": "He reports drinking 3 beers per day.",
      "label": "SOCIAL_HISTORY"
    },
    {
      "id": 5,
      "segment": "His temperature is 37.1°C and pulse is 78.",
      "shard": "Vital signs show a temperature of 37.1°C and a pulse of 78.",
      "label": "VITAL_SIGNS"
    },
    {
      "id": 6,
      "segment": "Which of the following is the most likely diagnosis?",
      "shard": "What is the most likely diagnosis given these findings?",
      "label": "QUESTION_STEM"
    }
  ]
}
\end{verbatim}

Now complete the task for the following question:

\begin{verbatim}
<question>
[[QUESTION]]
</question>

Output only valid JSON, no explanation.
\end{verbatim}

\end{promptbox}

\subsection{Turn-Granularity Repartitioning Prompt}
\begin{promptbox}[Turn-Granularity Repartitioning Prompt]
\ttfamily\small

Turn 1 (stem + all answer options) is fixed in the pipeline from raw data. You must only repartition the vignette below into exactly \texttt{[[N\_MINUS\_1]]} shards, which will become user turns 2..\texttt{[[N]]}.

Goal: produce exactly \texttt{[[N]]} total turns overall, counting the fixed Turn 1. You may rephrase the vignette for fluency, but you must preserve all source information faithfully.

\textcolor{red}{\textbf{Fixed turn 1}} (reference only --- do NOT output this; do NOT repeat or paraphrase stem/options in your shards):

\texttt{[[TURN1\_REFERENCE]]}

\textcolor{red}{\textbf{Vignette source}} --- split and rephrase ONLY this block:

\texttt{[[VIGNETTE\_SOURCE]]}

\textcolor{red}{\textbf{Instructions:}}
\begin{itemize}
\item Output exactly \texttt{[[N\_MINUS\_1]]} shards.
\item Preserve every clinical fact from the source, including all symptoms, negated findings, time expressions, quantities, measurements, lab values, exam findings, treatments, and chronology.
\item Do not omit, invent, generalize, or contradict any information.
\item Preserve the original order of information from the source; do not reorder facts across shards.
\item Prefer splitting at clinically natural boundaries (e.g., demographics/chief complaint, history of present illness, past history, exam, labs/imaging, treatment/course) rather than arbitrary sentence breaks.
\item Avoid trivial shards that contain only a fragment or isolated modifier unless necessary to satisfy the exact shard count.
\item Each shard should read naturally as a coherent user turn.
\end{itemize}

\textcolor{red}{\textbf{Hard rule:}}
\begin{itemize}
\item Do not include or paraphrase the diagnostic question.
\item Do not include or paraphrase any answer option text.
\item Do not include option letters such as (A), (B), (C), or (D).
\item If any sentence mixes clinical content with question/option wording, keep only the clinical content.
\end{itemize}

Do not use external knowledge beyond the vignette source.

Before finalizing, internally verify that all source facts are preserved across the shards and that the shard count is exactly \texttt{[[N\_MINUS\_1]]}.

\textcolor{red}{\textbf{Output format:}}

Return valid JSON only, with exactly one top-level key: \texttt{"vignette\_shards"}.

\begin{verbatim}
{
  "vignette_shards": [
    {"shard_id": 2, "shard": "..."},
    {"shard_id": 3, "shard": "..."}
  ]
}
\end{verbatim}

Requirements:
\begin{itemize}
\item Use exactly \texttt{[[N\_MINUS\_1]]} objects in the array.
\item Use \texttt{shard\_id} values 2, 3, ..., \texttt{[[N]]} in order.
\item Do not output any text before or after the JSON object.
\end{itemize}

\end{promptbox}

\subsection{Single-Turn Evaluation (Full / Concat)}
\begin{promptboxamber}[Single-Turn Baseline Prompt]
\ttfamily\small

You are a clinical expert. Given the following medical question, select the correct answer.

Return JSON only (no markdown, no extra text) with exactly this key:

\begin{verbatim}
{"answer":"<option letter or exact option text>"}
\end{verbatim}

Examples: \textcolor{red}{\texttt{\char123"answer":"A"\char125}}, \textcolor{red}{\texttt{\char123"answer":"(A) Androgenetic alopecia"\char125}}

No explanation, \textcolor{red}{\textbf{JSON only}}.

\end{promptboxamber}

\subsection{Ask-Question-First (Q-First)}
\begin{promptboxblue}[Ask-Question-First System Prompt]
\ttfamily\small

You are a clinical expert for multi-turn MedQA with question shown first.

At every turn, return JSON only (no markdown, no extra text) with exactly these keys:

\begin{verbatim}
{
  "action": "wait|answer|change",
  "answer": "<option letter or exact option text>",
  "confidence": <0.00-1.00>
}
\end{verbatim}

\textcolor{red}{\textbf{Rules:}}

\begin{enumerate}
\item If information is insufficient, set \textcolor{red}{\textbf{"action":"wait"}}.
\item On the first wait, \texttt{"answer"} can be empty: \texttt{""}.
\item If you have already provided an answer and now choose \texttt{"action":"wait"}, keep the same previous answer in \texttt{"answer"}.
\item Use \textcolor{red}{\textbf{"action":"answer"}} for your first answer.
\item Use \textcolor{red}{\textbf{"action":"change"}} when you want to change your mind; \texttt{"answer"} must be the new option.
\item \texttt{"confidence"} must always be a number between 0.00 and 1.00.
\item No explanation text, no additional keys, JSON only.
\end{enumerate}

\end{promptboxblue}

\subsection{Ask-Question-Last (Q-Last)}
\begin{promptboxteal}[Ask-Question-Last System Prompt]
\ttfamily\small

You are a clinical expert for multi-turn MedQA with question shown last.

At every turn, return JSON only (no markdown, no extra text) with exactly these keys:

\begin{verbatim}
{
  "action": "wait|answer|change",
  "answer": "<option letter or exact option text>",
  "confidence": <0.00-1.00>
}
\end{verbatim}

\textcolor{red}{\textbf{Rules:}}

\begin{enumerate}
\item If the message contains only clinical information (no question with options), set \textcolor{red}{\textbf{"action":"wait"}} and \texttt{"answer":""}.
\item If the message contains a medical question with answer options: use \textcolor{red}{\textbf{"action":"answer"}} for your first answer, or \textcolor{red}{\textbf{"action":"change"}} if you want to change your previous answer.
\item \texttt{"confidence"} must always be a number between 0.00 and 1.00.
\item No explanation text, no additional keys, JSON only.
\end{enumerate}

\end{promptboxteal}

\newpage

\end{document}